\documentclass[journal]{IEEEtran}
\usepackage[caption=false,font=normalsize,labelfont=sf,textfont=sf]{subfig}
\usepackage[pdftex]{graphicx}
\usepackage{stfloats}
\usepackage{rotating}
\usepackage{graphicx}  %Required
\usepackage{epstopdf}
\usepackage{amsmath}
\usepackage{booktabs}
\usepackage{amssymb}
\usepackage{textcomp,booktabs}
\usepackage{multirow}
\usepackage[usenames]{color}
\usepackage{booktabs}
\usepackage{makecell}
\usepackage{mathrsfs}
\usepackage{amsmath}
\usepackage{txfonts}
\usepackage{textcomp,booktabs}
\usepackage{multirow}
\usepackage{amsmath}
\usepackage[usenames]{color}
\usepackage{wrapfig}
\usepackage{algorithm}
\usepackage{algorithmic}
\usepackage{longtable}
\usepackage{colortbl,booktabs}
\usepackage{verbatim}
\usepackage[colorlinks,linkcolor=blue]{hyperref}

\begin{document}

\title{A Novel Long-term Iterative Mining Scheme for Video Salient Object Detection}


%\author{Anonymous}

%\author{Chenglizhao Chen$^{1*}$\thanks{Corresponding
%		author: Chenglizhao (cclz123@163.com)
%	~~~~Jia Song$^{1}$ ~~~~Yuming Fang$^{2}$~~~~Aimin Hao\textbf{$^{3}$} ~~~~Hong
%	Qin$^{3}$\\ $^1$Qingdao University ~~~~~$^2$ Jiangxi University of Finance and Economics
%	~~~~~~~~~~~~~~~$^3$Stony Brook University \\
%	\{cclz123, 15610452909, \}@163.com,  leo.fangyuming@foxmail.com, qin@cs.stonybrook.edu\\
%	Code\&Data: \url{https://github.com/qduOliver/MQP}\\
%%}

\author{Chenglizhao Chen$^{1}$
	\thanks{Corresponding author: Chong Peng (pchong1991@163.com)}
	~~~~Hengsen Wang$^{2}$~~~~Yuming Fang$^{3}$~~~~Chong Peng$^{1*}$\\
$^1$China University of Petroleum (East China)~~$^2$Qingdao University~~$^3$Jiangxi University of Finance and Economics\\
Code and Data: {\url{https://github.com/qduOliver/LIMVSOD}}\vspace{-0.8cm}\\
}

\markboth{IEEE Transactions on Circuits and Systems for Video Technology, VOL.XX, NO.XX, XXX.XXXX}%
{Shell \MakeLowercase{\textit{et al.}}: Bare Demo of IEEEtran.cls for Journals}

\maketitle

%\IEEEtitleabstractindextext{
\begin{abstract}
The existing state-of-the-art (SOTA) video salient object detection (VSOD) models have widely followed short-term methodology, which dynamically determines the balance between spatial and temporal saliency fusion by solely considering the current consecutive limited frames.
However, the short-term methodology has one critical limitation, which conflicts with the real mechanism of our visual system --- a typical long-term methodology.
As a result, failure cases keep showing up in the results of the current SOTA models, and the short-term methodology becomes the major technical bottleneck.
To solve this problem, this paper proposes a novel VSOD approach, which performs VSOD in a complete long-term way.
Our approach converts the sequential VSOD, a sequential task, to a data mining problem, i.e., decomposing the input video sequence to object proposals in advance and then mining salient object proposals as much as possible in an easy-to-hard way.
Since all object proposals are simultaneously available, the proposed approach is a complete long-term approach, which can alleviate some difficulties rooted in conventional short-term approaches.
In addition, we devised an online updating scheme that can grasp the most representative and trustworthy pattern profile of the salient objects, outputting framewise saliency maps with rich details and smoothing both spatially and temporally.
The proposed approach outperforms almost all SOTA models on five widely used benchmark datasets.
\end{abstract}

%}
\maketitle
%\IEEEdisplaynontitleabstractindextext
\IEEEpeerreviewmaketitle

\begin{figure}[t]
	\centering
	\includegraphics[width=1\linewidth]{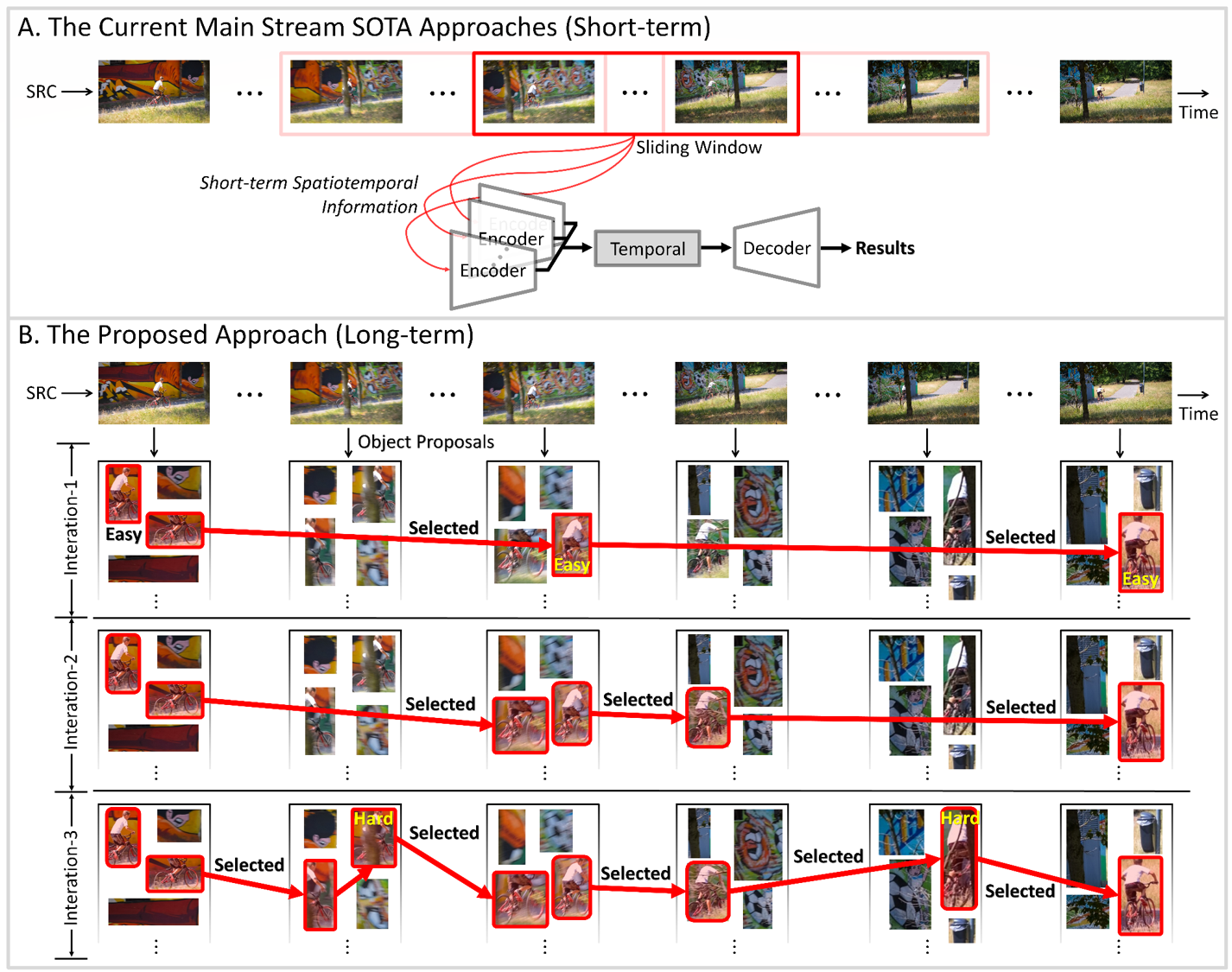}
    \vspace{-0.4cm}
	\caption{The major difference between the current SOTA models and our approach. As seen in subfigure-A, the conventional VSOD models usually take several consecutive video frames as input; thus, their VSOD methodology follows the short-term manner, where the current saliency decision is only derived on the current consecutive frames. In sharp contrast, our approach is based on object-level clustering, where all object proposals belonging to different frames (including the beyond-scope frames) are simultaneously available when making saliency prediction, and this is a typical long-term manner, where, compared with the short-term manner, the long-term methodology can be more robust when the salient objects have large appearances or movement changes. We use subfigure-B to visualize the salient object proposal mining process.\vspace{-0.4cm}}
	\label{fig:Motivation}
\end{figure}

\section{Introduction and Motivation}
The \textbf{v}ideo \textbf{s}alient \textbf{o}bject \textbf{d}etection (VSOD), also known as zero-shot video segmentation~\cite{wwg20ECCV,wang2015consistent,liu2021novel,huang2021novel,wang2019learning,wang2019inferring}, has received extensive research attention in recent years, whose primary objective is to segment video objects that attract the human visual attention most~\cite{chen17ST,chen2020videoTVCG,kompella2021semi}.
Different from the widely studied \textbf{i}mage \textbf{s}alient \textbf{o}bject \textbf{d}etection (ISOD) using spatial information only~\cite{han2018advanced,han2017unified,chen2020improved}, the temporal information provided by the video data makes the saliency detection task more difficult~\cite{cong2018review,kong2021self,xu2021multi}, and we give an in-depth discussion regarding this issue to clearly demonstrate our motivation.

Generally, compared with the spatial saliency cue, the human visual system is more easily attracted by the temporal saliency cue~\cite{WWG6,WWG4,bi2021sta}.
Taking an image with a complex scene for instance, we tend to focus on image regions that are spatially contrastive to their nearby surroundings in terms of colors or textures~\cite{chen2015structure,ma2019salient,chen2022video}.
However, our attention can promptly shift to another region if it involves sudden movement, even though this movement is very slight.
Thus, a preferable VSOD approach should not overemphasize temporal information because it is not always trustworthy; for example, the temporal saliency cue would be completely absent if the salient object stays static for a long period of time.
Therefore, the key for obtaining high-performance VSOD relies on how to balance the spatial and temporal saliency cues.

However, most of the current \textbf{s}tate-\textbf{o}f-\textbf{t}he-\textbf{a}rt (SOTA) deep models~\cite{fan2019shifting_ssav,li2019motion_mga,ren20TENet,tang2021video} have followed the short-term methodology, where the spatiotemporal balance is simply determined by the `current' multiple frames.
Actually, this short-term manner has one critical limitation:\\
Due to the varying nature of object movements, it is very difficult for the existing SOTA models~\cite{song2018pyramid_pdbm,chen2021confidence} to achieve an optimal spatiotemporal balance if only consecutive short-term frames are considered.
For example, from the short-term perspective, an object might be temporally salient in both spatial and temporal sources;
however, in the long-term view, it might be a nonsalient object.
In fact, the human visual system basically follows the long-term methodology because human beings tend to automatically remember the most representative appearance in the passing video contents.
Therefore, if a deep model follows the short-term methodology, it will easily produce failure cases when the current spatial and temporal saliency cues are less trustworthy.

Moreover, considering the VSOD task, we argue that the widely followed methodology - designing a very strong deep model to handle all cases - might be problematic and not necessary.
The main reason is that for video frames belonging to an identical video sequence, the spatial scene tends to be relatively stable; thus, even a very simple deep model can perform very well on these frames if it has been updated/fine-tuned online on the trustworthy saliency cues mined in an online manner.

Recently, there exist several works~\cite{Ref1TIP22,Ref2AAAI22} which have attempted to explore long-term information, yet our method is different with them in two key aspects.
First, both~\cite{Ref1TIP22,Ref2AAAI22} cannot make full use of long-term information, while our approach is global, where both the key frame selection and iteration processes follow a complete long-term way.
Second, compared with~\cite{Ref2AAAI22} in method design, our approach is clearly more comprehensive with elegant design in submodules.

Considering all issues mentioned above, this paper performs the VSOD task in a fully long-term manner (see Fig.~\ref{fig:Motivation}-B).
Given a video sequence, we use the off-the-shelf object detector~\cite{tan2020efficientdet} to acquire all potential object proposals in advance.
Thus, the conventional frame-by-frame VSOD task can be converted to a data mining problem: iteratively finding as many salient object proposals as possible in an easy-to-hard way.
Since all object proposals are simultaneously available during the mining process, our approach is clearly a typical long-term approach; thus, it does not need to consider the complicated spatiotemporal balance mentioned above.

In our approach, those clear salient object proposals - the easy ones - can be localized in the early mining iterations, then the consistency of these easy cases is learned and later used as the auxiliary knowledge to help find the harder cases, and this is fully detailed in Sec.~\ref{sec:SOPL}.
Once all salient object proposals have been mined, we use the existing ISOD model, which is online updated/fine-tuned on the salient object proposals, to output framewise saliency maps. The technical details regarding the online fine-tuning are detailed in Sec.~\ref{sec:FRVOFT}.
\begin{figure*}[t]
	\centering
	\includegraphics[width=1\linewidth]{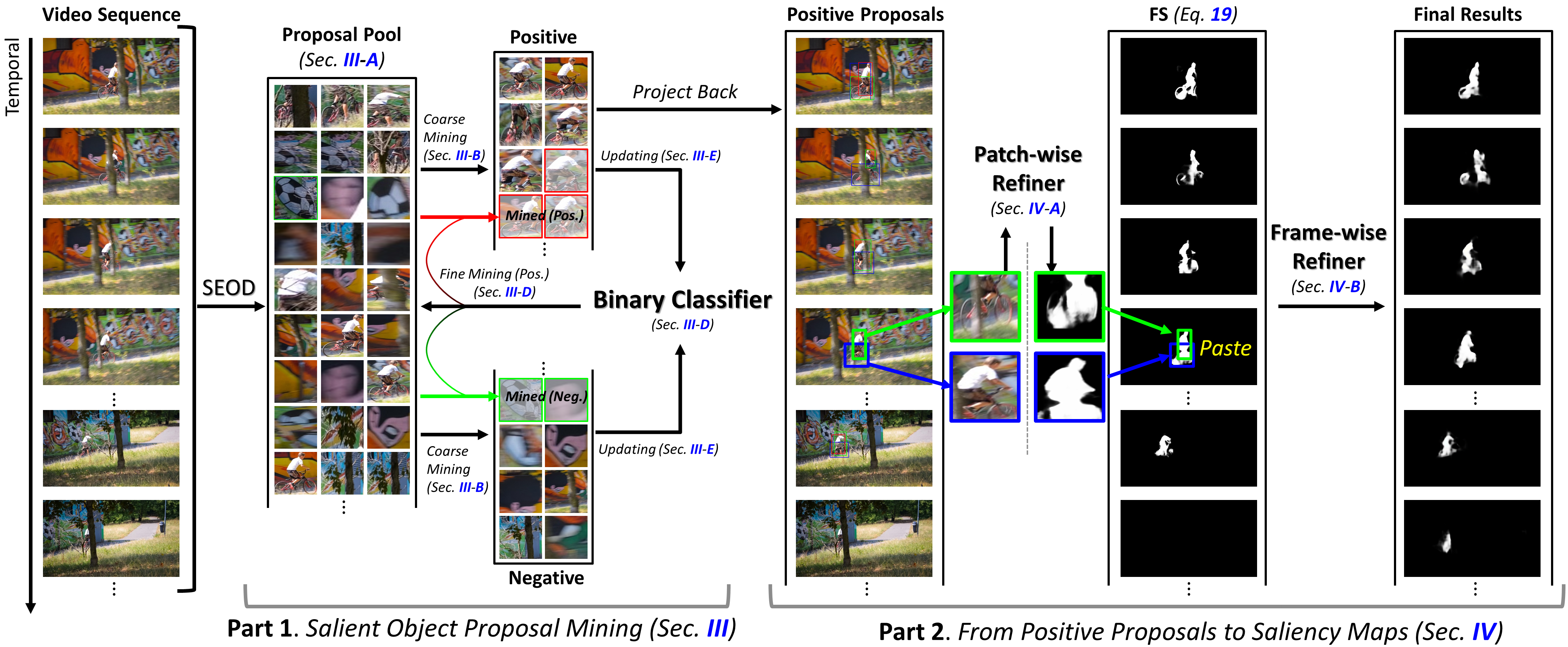}
    \vspace{-0.4cm}
	\caption{Method pipeline demonstration. SEOD: the off-the-shelf object decoder~\cite{tan2020efficientdet}. The proposed part 1 iteratively mines salient object proposals, and part 2 converts these salient object proposals into frame-wise saliency maps.\vspace{-0.4cm}}
	\label{fig:MethodPipeline}
\end{figure*}

The key contributions of this paper can be summarized into the following aspects:
\begin{itemize}
	\item
	We are one of the first attempts to convert the VSOD task, a conventional sequential task, into an orderless process, thus the long-term information can be well utilized to promote VSOD detection.
	\item
	We have devised a novel object proposal based mining scheme, which iteratively localizes salient objects in an easy-to-hard manner from a completely global perspective, thus it can avoid being trapped in considering the complicated spatiotemporal balance issue.
	\item
	A series of online saliency refinement strategies have been proposed, all of which ensure that our saliency inference process can take full advantage of the long-term mining process, producing accurate detection results with crisp object boundaries.
    \item
    A comprehensive quantitative evaluations and comparisons have been conducted to illustrate the effectiveness and superiority of the proposed approach.
    \item
Both codes and results are publicly available, which could be very potential to benefit our research community in the future.
\end{itemize}

\section{Related Work}
\subsection{Hand-crafted VSOD Approaches}
Although the main scope of this paper is deep learning-based VSOD, we give a brief review regarding the most representative conventional approaches.
Liu~\emph{et al}.~\cite{liuzhi16sp} performed the VSOD task via superpixel-based graph smoothing. By using color and global motion histograms, both spatial and temporal saliency cues are obtained in advance and then diffused via temporal propagation.
Xi~\emph{et al}.~\cite{xi16backgroundprior} devised a novel method for computing the background prior. This prior is combined by averaging the prior of the spatial~\cite{liu11siftflow} and temporal backgrounds~\cite{zhu14backp}.
Li~\emph{et al}.~\cite{li2018unsupervised} proposed a motion-based bilateral network to estimate the backgrounds in a framewise manner, which are later embedded into a graph for saliency propagation.
Chen~\emph{et al}.~\cite{chen2018scom} proposed a novel solution to distinguish moving salient objects from diverse changing background regions.
Zhou~\emph{et al}.~\cite{zhou18tmm} considered both temporal consistency and correlation among adjacent frames to compute the temporary saliency map. Then, both spatial and temporal saliency maps are fused via the proposed spatiotemporal refinement.
Guo~\emph{et al}.~\cite{guo2017video} designed a primitive approach to identify the salient object by ranking and selecting the salient proposals.
Chen~\emph{et al}.~\cite{chen17ST} adopted the low-rank strategy to alleviate the problem of false-alarm error accumulation induced by the widely used saliency propagation.
However, this approach basically follows the batchwise methodology, which fails to take full advantage of the beyond-scope saliency consistency, producing massive false alarms when the majority of the intrabatch saliency clues are incorrect.
To further improve, the same authors~\cite{chen2018bilevel} devised bilevel metric learning to include more spatial-temporal saliency cues in the current saliency prediction.
Guo \emph{et al}.~\cite{guo2019motion} proposed a fast VSOD method by using the principal motion vectors to represent the corresponding motion patterns, and such motion message coupling with the color clues together are fed into the multiclue optimization framework to achieve the spatiotemporal VSOD.
Zhao \emph{et al}.~\cite{zhao2021weakly} proposed a weakly-supervised VSOD model based on eye-fixation annotation. Compared with the fully-supervised  VSOD models, the proposed new annotation method dramatically reduces the consumption of time.

\subsection{Deep Learning-based VSOD Models}
In view of the video instance segmentation task~\cite{wang2021exploring,wang2022looking}, Liu \emph{et al.} in~\cite{VOSCVPR21} have presented a one-stage, end-to-end, and proposal-free deep model, whose key highlight is its capability of dividing instances into sub-regions dynamically and performing segmentation on each region for spatial granularity.
Thus, it achieves a more appealing segmentation behavior as it can enrich object details and produce masks with more accurate edges.
In view of the \textbf{v}ideo \textbf{o}bject \textbf{d}etection (VOD) task~\cite{lu2020learning}, Cui \emph{et al.} in~\cite{VOSICCV21} have proposed to depict the temporal feature relations and blend valuable neighboring features, thus the spatiotemporal feature representation can get enhanced. Also targeting at the VOD task, Liu \emph{et al.} in~\cite{NEUCOM2020VOS} proposed an approach which aggregates features calibrated at both pixel and instance levels, thereby achieving a better detection performance. More works about VOD tasks can be referred to the recent survey paper~\cite{wang2021survey}.

Let us now move to the most representative deep learning-based SOTA models, whose performances have significantly outperformed the conventional models.
Wang~\emph{et al}.~\cite{wang2017video_fcn} adopted an end-to-end network to compute spatial saliency, which was later coupled with two consecutive frames to serve as the input of another network to compute the spatiotemporal saliency.
Li~\emph{et al}.~\cite{li2019motion_mga} adopted the optical flow results as the network input to compute motion saliency.
Because motion saliency could only occasionally benefit the VSOD task, this work developed an attention mechanism, which is empowered by gate logic to filter the less trustworthy motion saliency cues when performing spatial and temporal saliency fusion.
Tokmakov~\emph{et al}.~\cite{tokmakov2017learning} fed the concatenated spatial and temporal deep features into the ConvLSTM for spatiotemporal saliency fusion.

Following the bistream structure, Le~\emph{et al}.~\cite{le2017deeply} computed the motion saliency via 3D convolutions, and the VSOD results were derived simply by concatenating and convolving the spatial and temporal deep features.
Chen~\emph{et al}.~\cite{chen2021_3d} proposed a new version of 3D convolution, which involves multiple 3D convolutions with the newly proposed temporal shuffle operation to enhance the network's ability in sensing temporal information.
This modification also enables the spatial branch to fully interact with the temporal branch in a multiscale manner.
Song~\emph{et al}.~\cite{song2018pyramid_pdbm} devised the bi-LSTM network to sense multiscale spatiotemporal information.
This work also adopted pyramid dilated convolutions to extract multiscale spatial saliency features, which are fed into the abovementioned bi-LSTM network to achieve multiscale VSOD.
Wang~\emph{et al}.~\cite{lu2019see} proposed coattention to enhance the consistency between different frames.
Fan \emph{et al}.~\cite{fan2019shifting_ssav} developed an attention-shift baseline and released a large-scale saliency-shift-aware dataset for the VSOD problem.
Zhou \emph{et al.}~\cite{ZhouTIP} have presented a novel end-to-end zero-shot video segmentation network. In this work, the proposed network follows the traditional bi-stream structure, yet, different from previous works, it newly devised a novel module to interact temporally with spatial.
Generally, this method follows a typical coarse-to-fine rationale.
The temporal information is treated as the coarse indicator to locate the target object. Then, based on the coarsely localized regions, the proposed fusion module further learns the spatial appearance to refine the segmentation process, ensuring the output has a crisp object boundary.
Recently, Zhou \emph{et al.}~\cite{ZhouCVPR} have devised a novel bottom-up instance discrimination network which takes advantage of temporal context information in videos for more accurate segmentation. In view of the methodology innovation, the authors have converted the video segmentation task to a tracking paradigm, which can better capture the appearance information of the target objects, yielding better segmentation results.

Recently, Gu~\emph{et al}.~\cite{gupyramid_pcsa} learned the nonlocal motion dependencies across several frames, and then it followed the pyramid structure to capture the spatiotemporal saliency clues at various scales.
The major highlight of this paper is the proposed constrained self-attention operation, which can capture motion cues via the prior that objects always move in a continuous trajectory, achieving a very high FPS rate.
Ren~\emph{et al}.~\cite{ren20TENet} proposed a novel triple-stream network, which includes a spatiotemporal network, a spatial network, and a temporal network, where the saliency cues computed from the spatial and temporal networks serve as the attention to enhance the main network.
Jiao~\emph{et al}.~\cite{jiao2021guidance} proposed a Guidance and Teaching Network (GTNet), GTNet introduces a temporal modulator to bridge features from motion into appearance, and a motion-guided mask is used to propagate the explicit cues during the feature aggregation.
Bi~\emph{et al}.~\cite{bi2021steg} devised STEG-Net, which uses the extracted edge cues to guide the extraction of spatial-temporal cues and combines deep texture cues with shallow edge cues. This strategy can simultaneously retain edge information and enhance object's global cues, making the object's position more accurately.

\subsection{Early Explorations Regarding Long-term Information}
Clearly, most of the current SOTA models mentioned above have one critical problem, i.e., their methodologies tend to be short-term in essence, and the limitations have been mentioned in the introduction.
Several previous works have become aware of this problem, and multiple modifications have been attempted to alleviate this problem.

Lu \emph{et al.} in~\cite{LuICCV19} have presented a GNN based \textbf{z}ero-shot \textbf{v}ideo \textbf{o}bject \textbf{s}egmentation (ZVOS) network. By using the newly devised \textbf{a}ttentive \textbf{g}raph \textbf{n}eural \textbf{n}etwork (AGNN), the ZVOS can be treated as an end-to-end message passing based graph information fusion procedure. The major highlight of this approach is its capability to make the high-order relations among frames be captured.
In the later journal version~\cite{LuTPAMI2}, the AGNN was extended to diverse segmentation tasks with additional in-depth discussions and explanations.
To further improve the SOTA models that primarily focus on learning discriminative foreground representations in a short-term manner, Lu \emph{et al.} in~\cite{LuTPAMI1} have proposed to re-design the ZVOS in a holistic fashion.
The authors have devised the co-attention layers to learn global correlations and scene context, and thus the short-term spatiotemporal feature fragments can be interacted in a joint feature space, achieving significant performance improvement.

Chen~\emph{et al}.~\cite{chen2019improved} used SIFT-Flow to introduce beyond-scope saliency cues to the current problem domain, making its spatiotemporal saliency computation relatively `long-term'.
Li~\emph{et al}.~\cite{li2019accurate} utilized tracking consistency to mine keyframes, and then the high-quality spatiotemporal saliency cues could be diffused from each keyframe to other normal frames.
The keyframe selection scheme proposed in this work belongs to the long-term scope, but the saliency diffusion process is still a short-term method.
Another representative work is~\cite{li2021tcsvt}, in which the keyframes were selected offline by measuring the consistency degree between spatial and temporal saliency cues, and then an ISOD model was fine-tuned on these keyframes to adapt for this sequence.
It is worth mentioning that this idea is inspired by~\cite{paul2017bmvc}, where the model fine-tuning scheme was used, while the major difference relies on whether the annotation of the 1st frame is given.
Wang~\emph{et al}.~\cite{Wang_2019_ICCV} proposed a novel attentive graph neural network to explore the relationship between different frames. Compared with the previous short-term method, the saliency sensing scope of this work was expanded significantly. However, because the network training and testing processes still followed the framewise sequential order, this work still belongs to the short-term category.
Following the model online matching and updating schemes that were widely used object trackers, Wang~\emph{et al}.~\cite{wwg20ECCV} proposed the episodic graph memory network, where multiple submodels are trained, stored, and updated to adapt the current VSOD task to the current video sequence long-term.
However, the VSOD task carried out by this work was solely along the temporal direction; thus, not all spatiotemporal saliency cues embedded in the video were simultaneously available to the current frame.
Wang~\emph{et al}.~\cite{wang2020improved} converted the framewise VSOD task to a graph problem, which converted the video frames to supervoxels, and the graph convolutional network, taking the supervoxels as the graph nodes, was used to explore the long-term spatiotemporal VSOD.
However, because this graph structure was still constrained by the temporal neighboring topology, the long-term information explored by this work might be rather weak in essence.
Zhang \emph{et al}.~\cite{zhang2021dynamic} proposed a dynamic context-sensitive filtering network (DCFNet). This net generates dynamic convolution kernels containing rich context information at multiple scales by estimating location correlation weights to improve the model's adaptability to dynamic video scenes.
Chen \emph{et al}.~\cite{chen2021novel} proposed a new concept, named motion quality, to re-balance the complementary fusion status between spatial information and temporal information.

\section{Salient Object Proposal Localization}
\label{sec:SOPL}
\subsection{Method Overview}
\label{sec:MO}
Given a video sequence, we use the off-the-shelf object detection tool SEOD~\cite{tan2020efficientdet} to acquire object proposals.
For a single frame, we rank all its object proposals according to their objectness confidences, and a maximum of 10 object proposals are considered (we also test to consider more than 10 object proposals, but no obvious performance improvement can be observed).
Thus, a video sequence with a total of $T$ frames can now be expressed as $N=T\times 10$ object proposals.
We use $P_i$ to denote the $i$-th object proposal, and all proposals can be represented as $\mathcal{P}=\{P_1...,P_N\}$.

The aim of this section is to find which object proposals are the salient proposals (see `Part 1' in Fig.~\ref{fig:MethodPipeline}).
Our key idea is to iteratively mine all potential salient object proposals from $\mathcal{P}$ from easy to hard, and in each mining iteration, only a small group of $\mathcal{P}$ could be determined to be either salient or nonsalient.
Initially, we place the accuracy rate as the highest priority, while the recall rate can be gradually ensured by the upcoming iterative mining steps.
The proposed iterative mining process can be visualized in Fig.~\ref{fig:Motivation}-B, where more salient object proposals, which are difficult to determine in the early iterations, can be correctly mined later because our mining approach can take full use of the knowledge derived from the previous iterations.

Once all salient object proposals have been determined, we utilize the online fine-tuning scheme to convert these salient object proposals to framewise saliency maps (see `Part 2' in Fig.~\ref{fig:MethodPipeline}).

\subsection{Coarse-level Localization for Salient Object Proposals}
\label{sec:CLFSOD}
For each object proposal in $\mathcal{P}$, we use the pretrained feature backbone (ResNet18) to extract the corresponding semantical deep feature ($f_i$), and the deep features of $\mathcal{P}$ can be represented as $\textbf{F}=\{f_1...,f_N\}$.

Based on the L2 similarity between each pair of \textbf{F}, all object proposals ($\mathcal{P}$) can be clustered into $K$ clusters.
Considering the limitation of GPU memory, long video sequences should be divided into multiple short sequences in advance.
It might also be rare for typical short video sequences (less than 100 frames) to contain more than three salient objects.
The clustering process can be represented as Eq.~\ref{eq:Clustering}.
\begin{equation}
\mathcal{C}=\{C_1...,C_K\}=Clustering(\textbf{F},K),
\label{eq:Clustering}
\end{equation}
where we use $C_i$ to denote the $i$-th cluster, $i\in\{1,2...,K\}$. Any off-the-self clustering tool can be applied here, where we simply choose the K-means.

To locate salient object proposals coarsely, the most intuitive method might be computing the `cluster-level saliency degree' first and then selecting the top salient object proposals as the salient clusters.
Our rationale is a salient cluster’s object proposals tend to exhibit very strong saliency cues.
Following this rationale, for each cluster, we use the averaged `motion saliency' to represent the cluster-level saliency degree.
Generally, compared with spatial saliency, motion saliency has one clear advantage: fewer false alarms; for example, at worst, motion saliency becomes absent, while spatial saliency might be totally incorrect for unpredictable reasons.
To continue introducing the technical details of the clustering-based salient object proposal coarse localization, we leave the contents regarding the motion saliency computation to the next subsection.

We select $\nu$ clusters from $\mathcal{C}$ as the salient clusters, where we empirically constrain $\nu<\lfloor 0.5\times K\rfloor$.
The exact value of $\nu$ can be automatically determined via Eq.~\ref{eq:sclsnum}.
\begin{equation}
\nu = arg\mathop{\min}_{i} \Big\{\breve{ams}(i)\Big\},
\label{eq:sclsnum}
\end{equation}
\begin{equation}
\breve{ams}\gets \nabla\big(\mathcal{DES}(ams)\big),
\label{eq:nabla}
\end{equation}
\begin{equation}
ams(i) = avg(PMS_i),
\label{eq:initialPMS}
\end{equation}
where $PMS_i=\{PMS_i(1)...,PMS_i(g)\}$ is a set in which each element represents the motion saliency value, e.g., $PMS_i(1)$ denotes the motion saliency value (1-dimension) of the 1st object proposal in the $i$-th cluster, and $g$ is the number of object proposals in this cluster; function $avg(C_i)$ returns the averaged motion saliency value of all object proposals in the $i$-th cluster, thus $ams$ is a $K$-dimensional vector; function $\mathcal{DES}(\cdot)$ sorts its input elements in descending order; $\nabla$ is the first-order derivation operation.

Clearly, the rationale of the dynamic $\nu$ relies on representing the initial `localization' regarding which object proposals among the cluster $C_i$ are very likely to be salient.

\subsection{Motion Saliency Cues}
\label{sec:MSC}
Intuitively, one can obtain motion saliency cues directly via an off-the-shelf \textbf{i}mage \textbf{s}alient \textbf{o}bject \textbf{d}etection (ISOD) model if it has been fine-tuned on the optical flow data, i.e., $\{OF_j, GT_j\}$ (see below).

\begin{figure}[t]
	\centering
	\includegraphics[width=1\linewidth]{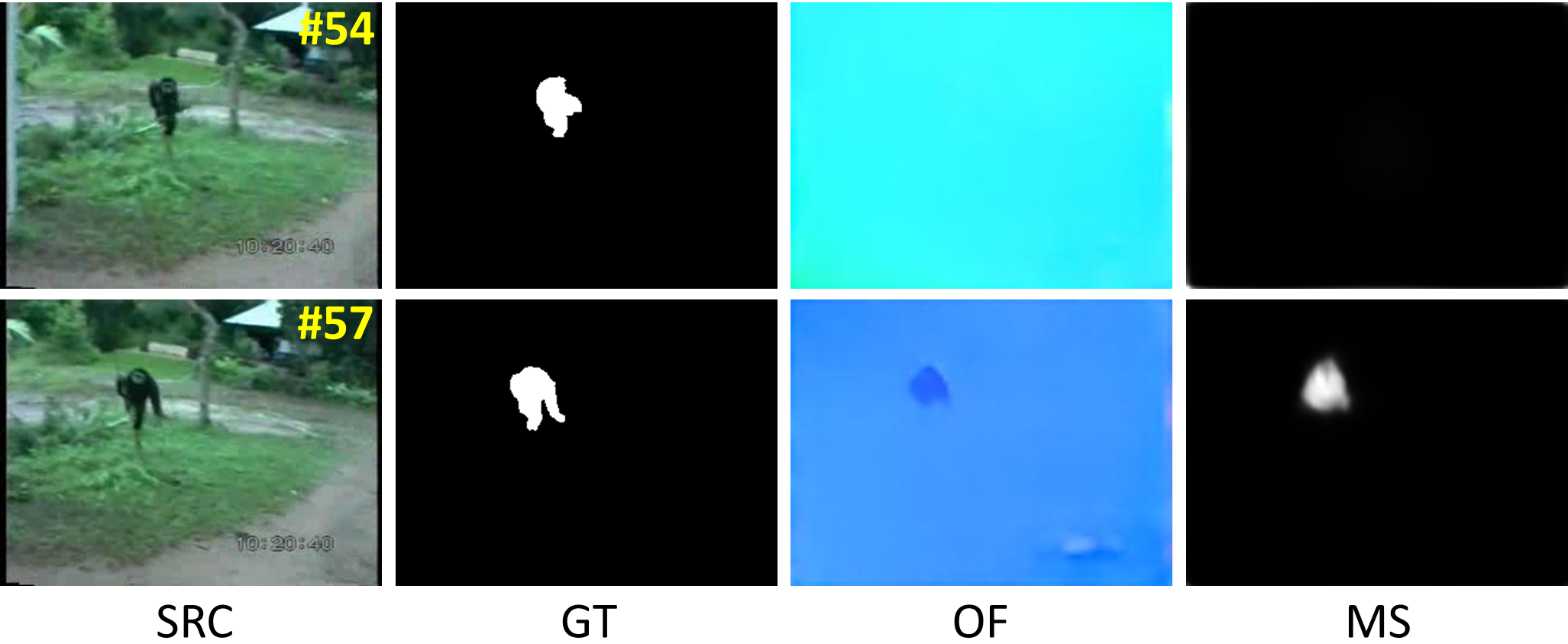}
    \vspace{-0.6cm}
	\caption{Pictorial demonstrations regarding the color encrypted optical flow results (OF) and the motion saliency maps (MS).}
%\vspace{-0.2cm}
	\label{fig:MS}
\end{figure}

Assume two consecutive video frames can be represented as $\textbf{I}_{j}$ and $\textbf{I}_{j+1}$, and the corresponding optical flow $OF_j$, which will be used for ISOD model fine-tuning, can be computed by Eq.~\ref{eq:OpticalFlow}.
\begin{equation}
OF_j = ce\Big\{FlowNet(\textbf{I}_j,\textbf{I}_{j+1})\Big\},
\label{eq:OpticalFlow}
\end{equation}
where `FlowNet' represents the off-the-shelf optical flow tool~\cite{sun2018pwc}, whose input includes two consecutive video frames, and the outputs are two gradient matrixes representing the spatial displacement over the vertical and horizontal directions; $ce\{\cdot\}$ denotes the widely used color encryption tool, which converts the abovementioned two optical flow matrixes to a three-dimensional matrix ($\mathbb{R}^{W\times H\times 3}$), in which all gradient directions and values are uniformly encoded by different colors after this operation (see the third column of Fig.~\ref{fig:MS}).

Thus, $OF_j$ and the corresponding raw ISOD ground truth $GT_j$ can be used to fine-tune the off-the-shelf ISOD model.
The loss function of ISOD deep model fine-tuning can be represented as Eq.~\ref{eq:ISODFinetune}, for which we use a small learning rate ($10^{-7}$).

\begin{equation}
\begin{split}
\rm L_{ISOD}=-\sum_\emph{j} \Big[\emph{GT}_\emph{j}&\times\log \emph{OF}_\emph{j}\\
&+(1-{\rm \emph{GT}}_\emph{j})\times\log(1-\emph{OF}_\emph{j})\Big].
\end{split}
\label{eq:ISODFinetune}
\end{equation}

After being fine-tuned, the adopted ISOD model ($\mathcal{M}$) takes $OF$ as input and then outputs a motion saliency map (see the last column in Fig.~\ref{fig:MS}). This process can be represented by the following equation:
\begin{equation}
\begin{split}
MS_j = \mathcal{M}(\Theta, OF_j),
\end{split}
\label{eq:MSCompute}
\end{equation}
where $\Theta$ is the learnable hidden parameters, $\mathcal{M}$ can be any off-the-shelf end-to-end ISOD model following the typical encoder-decoder structure (we choose the CPD19~\cite{wu2019cascaded_cpd}), and $MS_i$ is the motion saliency map (see the last column of Fig.~\ref{fig:MS}).

The motion saliency maps are usually blurred in object boundaries, which are mainly determined by the optical flow quality. To the best of our knowledge, there exists no better method for improving this problem at present.
Thus, the motion saliency maps can only be used for the coarse localization of the salient object proposals.

In addition, the motion saliency, a typical unstable saliency cue, tends to vary from frame to frame; for example, it may become helpless if the salient object is temporally static, as demonstrated in the \#54 frame in Fig.~\ref{fig:MS}.
However, the motion saliency cue might be very helpful in some other scenarios that have clear object movements, e.g., the \#57 frame.
Therefore, based on $MS$ (Eq.~\ref{eq:MSCompute}) and $\mathcal{P}$ (Sec.~\ref{sec:MO}), we detail the computation of $PMS$, which is mentioned in Eq.~\ref{eq:initialPMS}, as below.

Taking $PMS_i(u)$ as an example, suppose this object proposal is the $h$-th object proposal in $\mathcal{P}$ and it also belongs to the $j$-th video frame; thus, the exact value of $PMS_i(u)$ can be obtained by:
\begin{equation}
PMS_i(u)=avg\big(crop(MS_j, P_h)\big),
\label{eq:PMS}
\end{equation}
where the $crop$ operation crops a patch from $MS_j$ according to the coordinates provided by the object proposal $P_h$.

\subsection{Fine-level Localization for Salient Object Proposals}
\label{sec:FLSOP}
By performing the previous steps (e.g., Eq.~\ref{eq:sclsnum}), $\nu$ salient clusters have been selected.
The remaining problem here includes two aspects: 1) these $\nu$ salient clusters might contain some nonsalient object proposals, and 2) some salient object proposals might belong to the $
\{K-\nu\}$ nonsalient clusters.
To precisely separate these two types of object proposals (we call them the `undetermined' proposals), we resort to an auxiliary binary classifier.

Our idea is that for the `undetermined' object proposals, the binary classifier serves as an indicator to tell which object proposals are more likely to be salient.
In practice, this binary classifier can be weakly trained by instances belonging to salient clusters or nonsalient clusters.
The main reason that we call this training process a `weakly supervised' process is that for each video sequence, both positive and negative instances for the binary classifier training are determined online.
Without knowing the ground truths, some of the most trustworthy object proposals in the salient clusters are selected as the positive instances ($PseudoGT=1$).
The most trustworthy negative instances ($PseudoGT=0$) are selected from the nonsalient clusters.
To focus on the main topic in this subsection, we leave the technical details regarding instance selection to the next subsection.

Based on the positive and negative instances mentioned above (total $Q$ instances and $Q\le N$), the loss function of the binary classifier, a typical cross-entropy loss, can be represented as Eq.~\ref{eq:classloss}.
\begin{equation}
\begin{split}
\mathcal{L}_{BC}=\frac{1}{Q}\times\sum_q &-\Big[PseudoGT_q\times log\big(FC(f_q)\big)\\
&+(1-PseudoGT_q)\times log\big(1-FC(f_q)\big)\Big],
\end{split}
\label{eq:classloss}
\end{equation}
where $FC(\cdot)$ denotes the widely used multilayer fully connected layers, $f_q$ represents the high-dimensional deep feature of the $q$-th object proposal selected under the current salient and nonsalient cluster partition,
$PseudoGT_q$ is the corresponding binary label of $f_q$.
Note that the training process of this binary classifier of this classifier can be performed very quickly; thus, it can be applied in the VSOD task.
In the next subsection, we detail the computational process of $PseudoGT$.

\subsection{Training Instances for the Binary Classifier}
Considering that the performance of the binary classifier is positively related to the quality of the pseudolabels (i.e., $PseudoGT$), we should ensure that those less trustworthy object proposals are excluded from the pseudotraining set.
To achieve this, we utilize two measurements: 1) the distance to the cluster's centroid and 2) the motion saliency degree of each object proposal.

The rationale of measurement 1) is that the trustworthy degree of an object proposal to be a real salient proposal should be positively related to the confidence degree of belonging to its cluster.
Actually, this confidence degree can be measured by the feature similarity degree ($sim$) between the cluster's average profile ($AP_i$) and the object proposal (see Eq.~\ref{eq:sim}), where $AP_i$ could be automatically obtained during the K-means clustering process.
\begin{equation}
sim_i(j)=\big|\big|AP_i,f_j\big|\big|_2,\ \ AP_i = \frac{1}{g}\sum_{l=1}^g f_l,
\label{eq:sim}
\end{equation}
where $f_j$ represents the deep feature of the $j$-th object proposal in its cluster $C_i$, $||\cdot||_2$ is the $L_2$-norm, and $AP_i$ represents the clustering profile of $C_i$, which shares an identical size to the $f_j$; $sim_i(j)$ measures the feature distance of the $j$-th object proposal to the cluster centroid.
The primary objective of Eq.~\ref{eq:sim} is to serve the proposed mining process to filter out those less trustworthy object proposals in a given cluster.
Our motivation is clear and intuitive, \emph{i.e.}, given a salient cluster, an object belonging to yet with a large feature distance to the cluster's average profile tends to be non-salient. Thus, this object proposal shall have a large chance of being filtered during the subsequent mining process.
Once $sim$ (Eq.~\ref{eq:sim}) is obtained, we use it to rank all object proposals in ascending order.

%One can simply resort to the off-the-shelf RPCA (\textbf{r}obust \textbf{p}rinciple \textbf{c}omponent \textbf{a}nalysis)~\cite{Peng19rpca} to obtain $LP$, which can be detailed as Eq.~\ref{eq:LK}.
%\begin{equation}
%\min\limits_{LP,E} \big|\big|LP\big|\big|_* + \lambda\cdot\big|\big|E\big|\big|_1, \ \ s.t., F = LP+E,
%\label{eq:LK}
%\end{equation}
%where $F=\{f_1,f_2,...,f_g\}$, and $g$ represents the total number of object proposals in the current cluster; $E$ represents the sparse part, sharing the same size to $F$; $||\cdot||_*$ denotes the nuclear norm, and $||\cdot||_1$ is the $L_1$-norm; $\lambda$ is a parameter to balance the low-rank part ($LP$) and the sparse part ($E$).
%In our implementation, we simply follow the default setting provided by the RPCA tool.
%
%The low-rank computation process mentioned above will be performed independently for each cluster, and the initial rank number can be assigned to 1 to accelerate the convergency speed, which could only retain the largest eigen-value when performing the eigen-thresholding for pursuing the nuclear part.
%Under this strong constraint, the low-rank computation process is extremely fast.
%A more grand design regarding the low-rank part may achieve to some additional performance gain, but it seems to beyond the main scope of this paper, which deserves for some future investigation.

The rationale for measurement 2) relies on the fact that object proposals with clear movements are more likely to be salient proposals.
Thus, all object proposals in each cluster can also be ranked in descending order according to their motion saliency degrees, i.e., $PMS$ (Eq.~\ref{eq:PMS}).

Based on the abovementioned factors, the less trustworthy object proposals in the salient clusters ($\mathcal{C}^+$) or the nonsalient clusters ($\mathcal{C}^-$) can be filtered by using the equations documented below; $\mathcal{C}=\mathcal{C}^+\cup\mathcal{C}^-$ and $\cup$ is the union operation.
\begin{equation}
SI_i^+=\mathcal{ASD}\big(C_i^+, SIM_i\big),\ \ MI_i^+=\mathcal{DES}\big(C_i^+, PMS_i\big),
\label{eq:ASDDEC}
\end{equation}
\begin{equation}
SI_i^-=\mathcal{ASD}\big(C_i^-, SIM_i\big),\ \ MI_i^-=\mathcal{ASD}\big(C_i^-, PMS_i\big),
\label{eq:ASDDEC2}
\end{equation}
where we use the superscripts $+$ and $-$ to distinguish salient clusters and nonsalient clusters determined by the previous steps; $SIM_i\in\mathbb{R}^{1\times g}=[sim_i(1)..., sim_i(g)]$, and $sim_i(1)$ denotes the similarity degree of the 1st object proposal in the $i$-th cluster computed via Eq.~\ref{eq:sim}; $PMS_i$ can be found in Eq.~\ref{eq:PMS};
$SI_i, MI_i\in\mathbb{R}^{1\times g}$ are two subscript vectors of object proposals in the $i$-th salient cluster;
$\mathcal{ASD}(C_i, SIM_i)$ returns the reranked object proposals' subscripts via $sim_i$ in ascending order, and $\mathcal{DES}(C_i, MS_i)$ returns the reranked object proposals' subscripts via $MS_i$ in descending order.

Next, the intersection between the top-$\alpha$ elements in $SI_i$ and the top-$\beta$ in $MI_i$ can be the most trustworthy object proposals in the current cluster, where $\alpha$ and $\beta$ can be obtained via Eq.~\ref{eq:asdb} and Eq.~\ref{eq:decc}.
\begin{equation}
\alpha^+_i = arg\max_{\xi \leq j < g} \Big\{\nabla\big(Re(SIM_i,\ SI_i^+)\big)\Big\}\big(j
\big),
\label{eq:asdb}
\end{equation}
\begin{equation}
\alpha^-_i = arg\max_{\xi \leq j < g} \Big\{\nabla\big(Re(SIM_i,\ SI_i^-)\big)\Big\}\big(j
\big),
\label{eq:asde}
\end{equation}
\begin{equation}
\beta^+_i = arg\max_{\xi \leq j < g} \Big\{\nabla\big(Re(PMS_i,\ MI_i^+)\big)\Big\}\big(j
\big),
\label{eq:decc}
\end{equation}
\begin{equation}
\beta^-_i = arg\max_{\xi \leq j < g} \Big\{\nabla\big(Re(PMS_i,\ MI_i^-)\big)\Big\}\big(j
\big),
\label{eq:decf}
\end{equation}
where the definitions of $SI$ and $MI$ can be found in Eq.~\ref{eq:ASDDEC} and Eq.~\ref{eq:ASDDEC2},
the operation $Re(SIM_i,\ SI_i^+)$ reorders the elements in $SIM_i$ according to the $SI_i^+$,
similar to Eq.~\ref{eq:nabla}, $\nabla(\cdot)$ computes the gradient of its input vector, i.e., the output of $Re(\cdot,\cdot)$,
$\xi$ is the lower bound, and we empirically assign it to $0.6\times g$, and $g$ is the total proposal number in the $i$-th cluster.

The filtering process outputting trustworthy positive and negative instances can be detailed as Eq.~\ref{eq:posfil}.
\begin{equation}
\begin{split}
Pos\gets \Bigg\{TOP\Big(\mathcal{ASD}(&C_i^+,SIM_i),\alpha^+_i\Big)\Bigg\}\\
&\cap\ \Bigg\{TOP\Big(\mathcal{DES}(PMS_i),\beta_i^+\Big)\Bigg\},
\label{eq:posfil}
\end{split}
\end{equation}
\begin{equation}
\begin{split}
Neg\gets \Bigg\{TOP\Big(\mathcal{ASD}(&C_i^-,SIM_i),\alpha^-_i\Big)\Bigg\}\\
&\cap\ \Bigg\{TOP\Big(\mathcal{ASD}(PMS_i),\beta_i^-\Big)\Bigg\},
\label{eq:negfil}
\end{split}
\end{equation}
where $\mathcal{ASD}/\mathcal{DES}$ is identical to that of the previous equations; $\alpha/\beta$ can be found in Eq.~\ref{eq:asdb}, Eq.~\ref{eq:asde}, Eq.~\ref{eq:decc}, and Eq.~\ref{eq:decf};
$TOP(SIM_i,\alpha)$ returns the top-$\alpha$ elements of $SIM_i$, $SIM/PMS$ can be referred from Eq.~\ref{eq:ASDDEC}, and $\cap$ is the operation that returns the intersection of two sets.
Compared with the original $\mathcal{P}^+$ and $\mathcal{P}^-$, samples in $Pos$ or $Neg$ are clearly more trustworthy in general, and the total sample numbers of $Pos$ and $Neg$ should be less than the total object proposal number ($N$) of $\mathcal{P}$.

\begin{figure}[t]
	\centering
	\includegraphics[width=0.8\linewidth]{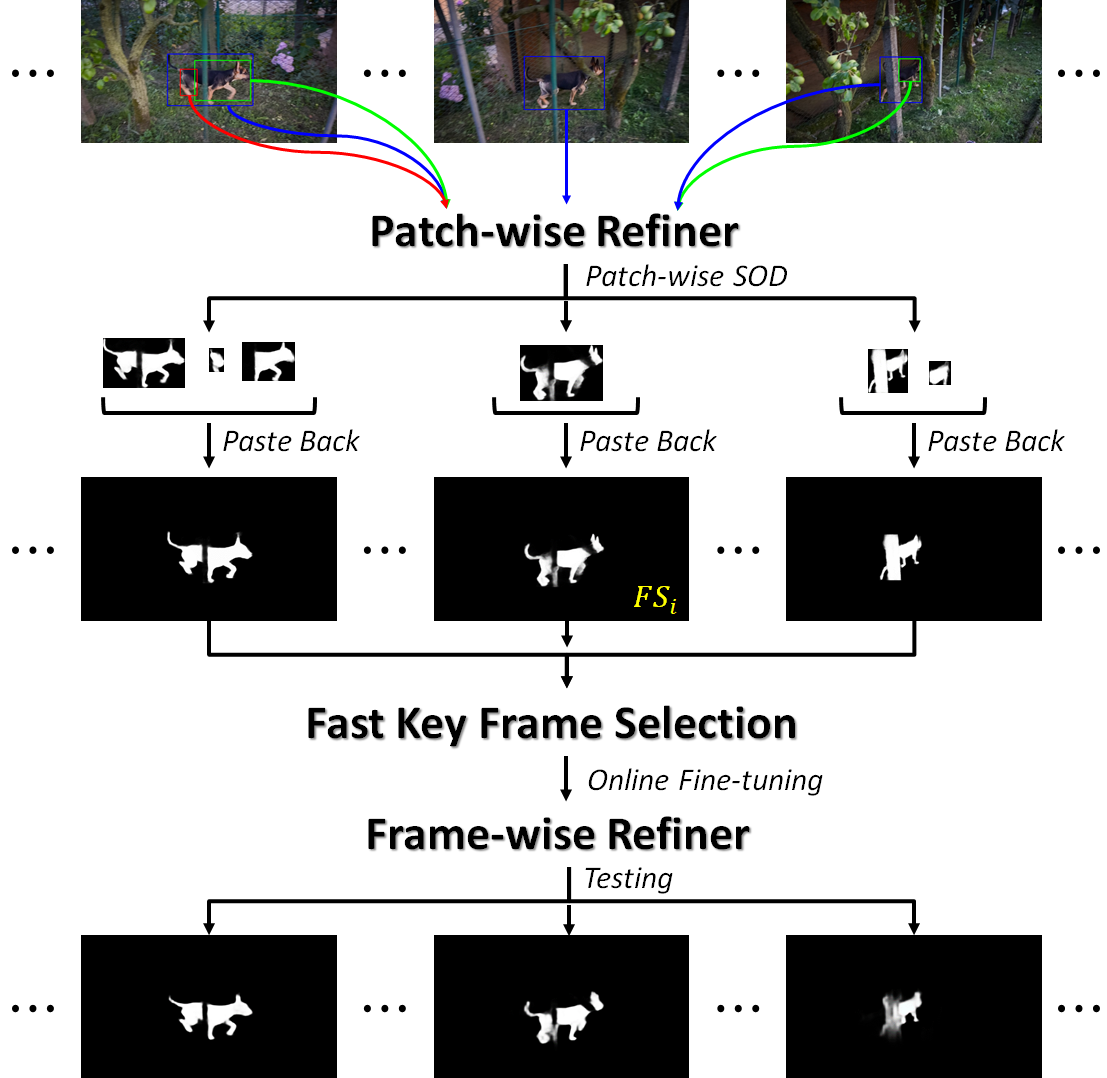}
    \vspace{-0.2cm}
	\caption{The detailed pipeline of the framewise refinement (Sec.~\ref{sec:FRVOFT}).}
    \vspace{-0.4cm}
	\label{fig:Online}
\end{figure}

\subsection{Iteratively Mining More Trustworthy Object Proposals}
\label{sec:IMMTOP}
Based on both $Pos$ and $Neg$, the binary classifier can be well trained.
By using this classifier as the indicator, we can determine the saliency labels (salient or nonsalient) for more object proposals that are relatively less trustworthy than those belonging to either $Pos$ or $Neg$, and we call them the `uncertain' object proposals.

For each uncertain object proposal in the salient clusters ($\mathcal{C}^+$), we add it into the $Pos$ set if the binary classifier predicts it as `salient'.
Similarly, for each uncertain object proposal in the nonsalient clusters ($\mathcal{C}^-$), we add it to the $Neg$ set if the binary classifier predicts it to be `nonsalient'.
By using the mining step mentioned above, both sets ($Pos$ and $Neg$) can be gradually expanded to comprise more feature patterns.
However, the binary classifier is not always trustworthy and predictions might be occasionally incorrect, making both $Pos$ and $Neg$ sets noisy.

To alleviate this problem, we propose two additional constraints to ensure that the uncertain object proposals included in $Pos$ and $Neg$ sets are the most trustworthy.
We use $Pos+$ and $Neg+$ to denote these two types of uncertain proposals.
Thus, only those object proposals, which belong to the $Pos+$ or $Neg+$ set and meet the constraint addressed below, would be eventually added into the $Pos$ or $Neg$ set.
Taking $Pos+$ as an example, all proposals are reordered via their similarity degrees to the mean profile of its cluster (Eq.~\ref{eq:sim}); thus, we denote the reordered $Pos+$ as $Pos+'$.
We constrain only the top-$\gamma\%$ proposals in $Pos+'$ to be added to the $Pos$ set.
We empirically choose a relative slack value (60\%) to $\gamma$ to balance the diversity and the trustworthiness degree.

\begin{figure}[t]
	\centering
	\includegraphics[width=1\linewidth]{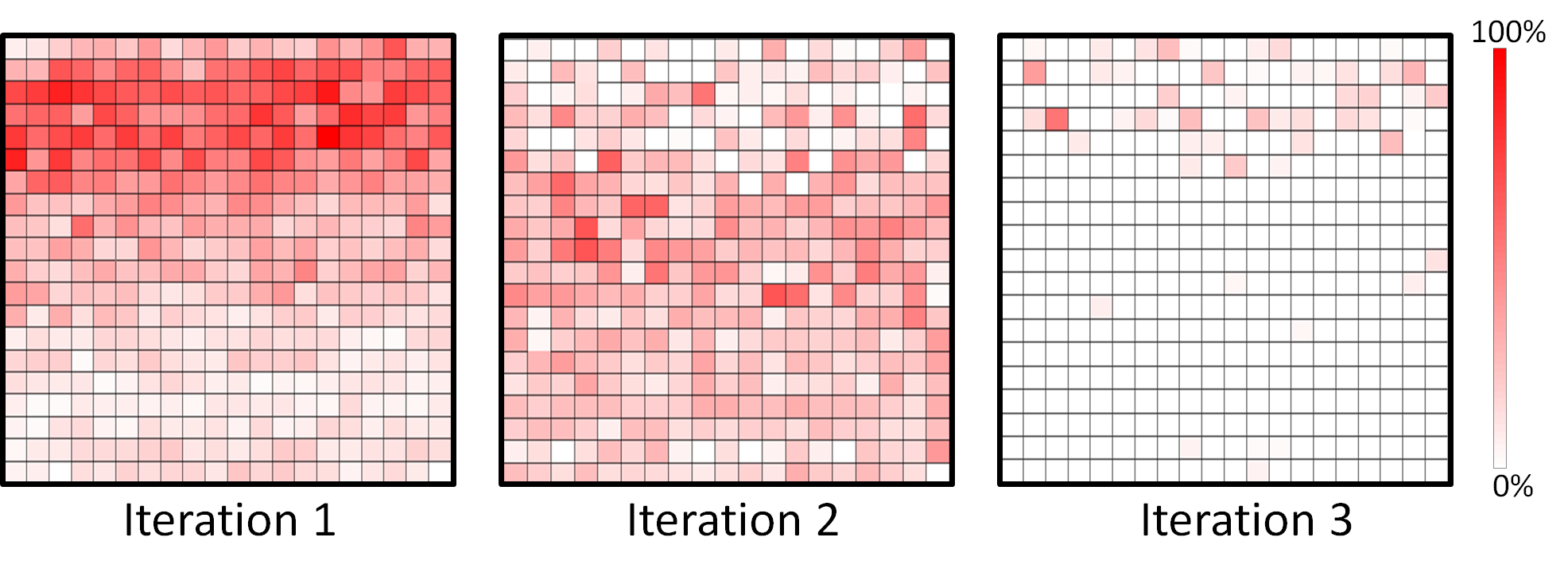}
    \vspace{-0.8cm}
	\caption{Visualization of the iterative data mining process. For each video sequence in the Davis dataset, all its object proposals are ordered according to their `trustworthy degrees'. We measure the trustworthiness degree via the average number of nonzero ground truth pixels inside the object proposal. Thus, all object proposals in a video sequence can be ordered in descending order according to the trustworthy degree, which can be represented in the form of a 1-dimensional vector. Since the object proposal numbers tend to vary with different video sequences, we resize this 1-dimensional vector to a fixed size (400). Then, each 1-dimensional vector is reshaped to a 2-dimensional matrix. In each iteration, each object proposal can reach two statuses: being selected (1) or not (0). Thus, the mining status of each video sequence can be visualized as a binary matrix, where all its elements belong to 1 or 0. We average all these binary matrixes of the Davis dataset as one probability matrix to show the probability of object proposals with different trustworthy degrees to be selected in each mining iteration.}
    \vspace{-0.4cm}
	\label{fig:Iteration}
\end{figure}

In our implementation, we repeat the above mining process multiple times, where both $Pos$ and $Neg$ can be expanded each time.
Specifically, to acquire more knowledge, the binary classifier is updated/fine-tuned after both $Pos$ and $Neg$ are expanded.
Clearly, the binary classifier becomes more powerful as the above mining process continues.

We visualize the iterative mining process regarding the salient object proposals in Fig.~\ref{fig:Iteration}.
As seen in `iteration 1', the mining process tends to simply select the most trustworthy object proposals (the top-left cell is the most trustworthy object proposal, and the bottom-right cell is the cell with the smallest trustworthy degree).
Then, in `iteration 2', some of the less trustworthy object proposals are selected to enhance the diversity of the $Pos$ set.
Finally, in `iteration 3', again, some of the most trustworthy object proposals that are misdetected by the previous iterations can be selected.
Compared with the previous two iterations, only a very small group of object proposals are selected in `iteration 3'; thus we omit `iteration 4' to ensure efficiency.

\section{Framewise Refinement via Online Fine-tuning}
\label{sec:FRVOFT}
Considering that the object proposals belonging to the $Pos$ set (Sec.~\ref{sec:IMMTOP}) are very likely to be the real salient proposals, we propose fine-tuning the existing ISOD model online on the knowledge embedded in the `$Pos$' set.
Thus, this model, which has shown some clips of the salient objects, can segment the salient objects well in the current video sequence.
The proposed online model fine-tuning is demonstrated in Fig.~\ref{fig:Online}, which includes three major components: 1) patchwise refiner, 2) fast key frame selection, and 3) framewise refiner.

\subsection{Patchwise Refiner}
\label{sec:PWR}
Previous steps (Sec.~\ref{sec:SOPL}) can only tell us which object proposals are salient.
To obtain a framewise saliency map, we simply resort to the existing \textbf{i}mage \textbf{s}alient \textbf{o}bject \textbf{d}etection (ISOD) model, i.e., each object proposal is fed into the ISOD model to obtain a patchwise saliency map.

\begin{table*}[t]
	\centering
    \Huge
	\caption{Quantitative comparisons with current SOTA methods. The top three results are marked by red, green and blue, respectively.}
\vspace{-0.2cm}
	\renewcommand\arraystretch{1.2}
	\resizebox{1\textwidth}{!}{
		\begin{tabular}{|r|c||c|cc|cccc|c|c|c|c|c|c|c|c|}
			\hline
			\multirow{3}[1]{*}{Dataset} & \multirow{3}[1]{*}{Metrics} & \multirow{3}[1]{*}{Ours}  & \multicolumn{2}{c|}{2021}& \multicolumn{4}{c|}{2020}     & \multicolumn{4}{c|}{2019}     & \multicolumn{3}{c|}{2018} & \multicolumn{1}{c|}{2017} \\
			&       &      &DCFNet\cite{zhang2021dynamic} &MQP\cite{chen2021novel} & TENet\cite{ren20TENet} & U2Net\cite{qin2020u2} & PCSA\cite{gupyramid_pcsa} & LSTI\cite{chen2019improved}  & \multicolumn{1}{c}{SSAV\cite{fan2019shifting_ssav}} & \multicolumn{1}{c}{MGA\cite{li2019motion_mga}} & \multicolumn{1}{c}{COS\cite{lu2019see}} & CPD\cite{wu2019cascaded_cpd}   & \multicolumn{1}{c}{PDB\cite{song2018pyramid_pdbm}} & \multicolumn{1}{c}{MBN\cite{li2018unsupervised}} & SCO\cite{chen2018scom}  & \multicolumn{1}{c|}{SFLR\cite{chen17ST}} \\
			\hline
		
			\hline
			& maxF  & \textcolor[rgb]{1.000, 0.000, 0.000}{\textbf{0.911}} &\textcolor[rgb]{0.000, 0.439, 0.753}{\textbf{0.900}} &{\textcolor[rgb]{0.000, 0.690, 0.314}{\textbf{0.904}}} & 0.881 & 0.839  & 0.880  & 0.850  & \multicolumn{1}{c}{0.861} & \multicolumn{1}{c}{0.892} & \multicolumn{1}{c}{0.875} & 0.778  & \multicolumn{1}{c}{0.855} & \multicolumn{1}{c}{0.861} & 0.783  & \multicolumn{1}{c|}{0.727}   \\
			Davis\cite{perazzi2016benchmark} & S-M   & \textcolor[rgb]{1.000, 0.000, 0.000}{\textbf{0.922}}&\textcolor[rgb]{0.000, 0.439, 0.753}{\textbf{0.914}}&{\textcolor[rgb]{0.000, 0.690, 0.314}{\textbf{0.916}}} &0.905 & 0.876  & 0.902  & 0.876  & \multicolumn{1}{c}{0.893} & \multicolumn{1}{c}{0.910} & \multicolumn{1}{c}{0.902} & 0.859  & \multicolumn{1}{c}{0.882} & \multicolumn{1}{c}{0.887} & 0.832  &\multicolumn{1}{c|}{0.790}   \\
			& MAE   & \textcolor[rgb]{1.000, 0.000, 0.000}{\textbf{0.016}}&\textcolor[rgb]{1.000, 0.000, 0.000}{\textbf{0.016}}&{\textcolor[rgb]{0.000, 0.439, 0.753}{\textbf{0.018}}}& \textcolor[rgb]{0.000, 0.690, 0.314}{\textbf{0.017}} & 0.027  & 0.022  & 0.034  & \multicolumn{1}{c}{0.023} & \multicolumn{1}{c}{0.023} & \multicolumn{1}{c}{0.020} & 0.032  & \multicolumn{1}{c}{0.028} & \multicolumn{1}{c}{0.031} & 0.064  & \multicolumn{1}{c|}{0.056}  \\
		
			\hline
			& maxF  & \textcolor[rgb]{1.000, 0.000, 0.000}{\textbf{0.899}}&0.839& {\textcolor[rgb]{0.000, 0.439, 0.753}{\textbf{0.841}}}& 0.810  & 0.775  & 0.810  & \textcolor[rgb]{0.000, 0.690, 0.314}{\textbf{0.858}} & \multicolumn{1}{c}{0.801 } & \multicolumn{1}{c}{0.821} & \multicolumn{1}{c}{0.801} & 0.778  & \multicolumn{1}{c}{0.800} & \multicolumn{1}{c}{0.716} & 0.764  & \multicolumn{1}{c|}{0.745}   \\
			Segv2\cite{li2013video} & S-M   & \textcolor[rgb]{1.000, 0.000, 0.000}{\textbf{0.921}} & \textcolor[rgb]{0.000, 0.690, 0.314}{\textbf{0.883}}&\textcolor[rgb]{0.000, 0.439, 0.753}{\textbf{0.882}} & 0.868 & 0.843  & 0.865  &0.870 & \multicolumn{1}{c}{0.851} & \multicolumn{1}{c}{0.865} & \multicolumn{1}{c}{0.850} & 0.841  & \multicolumn{1}{c}{0.864} & \multicolumn{1}{c}{0.809} & 0.815  & \multicolumn{1}{c|}{0.804}   \\
			& MAE   & \textcolor[rgb]{1.000, 0.000, 0.000}{\textbf{0.013}}&{\textcolor[rgb]{0.000, 0.690, 0.314}{\textbf{0.015}}}&{\textcolor[rgb]{0.000, 0.439, 0.753}{\textbf{0.018}}}& 0.025  & 0.042  & 0.025  & 0.025  & \multicolumn{1}{c}{0.023} & \multicolumn{1}{c}{0.030} & \multicolumn{1}{c}{0.020} & 0.023  & \multicolumn{1}{c}{0.024 } & \multicolumn{1}{c}{0.026} & 0.030  & \multicolumn{1}{c|}{0.037} \\
			
			\hline
			& maxF  & \textcolor[rgb]{0.000, 0.439, 0.753}{\textbf{0.953}}&\textcolor[rgb]{0.000, 0.439, 0.753}{\textbf{0.953}}& 0.939& 0.949  & \textcolor[rgb]{0.000, 0.690, 0.314}{\textbf{0.958}} & 0.940  & 0.905  & \multicolumn{1}{c}{0.939} & \multicolumn{1}{c}{0.933} & \multicolumn{1}{c}{\textcolor[rgb]{1.000, 0.000, 0.000}{\textbf{0.966}}} & 0.941  & \multicolumn{1}{c}{0.888} & \multicolumn{1}{c}{0.883} & 0.831  & \multicolumn{1}{c|}{0.779}   \\
			Visal\cite{wang2015consistent} & S-M   & 0.947& \textcolor[rgb]{0.000, 0.690, 0.314}{\textbf{0.952}}& 0.942 & \textcolor[rgb]{0.000, 0.439, 0.753}{\textbf{0.949}} & \textcolor[rgb]{0.000, 0.690, 0.314}{\textbf{0.952}} & 0.946  & 0.916  & \multicolumn{1}{c}{0.943} & \multicolumn{1}{c}{0.936} & \multicolumn{1}{c}{\textcolor[rgb]{1.000, 0.000, 0.000}{\textbf{0.965}}} & 0.942  & \multicolumn{1}{c}{0.907} & \multicolumn{1}{c}{0.898} & 0.762  &\multicolumn{1}{c|}{0.814}   \\
			& MAE   & \textcolor[rgb]{0.000, 0.690, 0.314}{\textbf{0.011}}&\textcolor[rgb]{1.000, 0.000, 0.000}{\textbf{0.010}}&0.016 & \textcolor[rgb]{0.000, 0.439, 0.753}{\textbf{0.012}} & \textcolor[rgb]{0.000, 0.690, 0.314}{\textbf{0.011}} & 0.017  & 0.033  & \multicolumn{1}{c}{0.020} & \multicolumn{1}{c}{0.017} & \multicolumn{1}{c}{\textcolor[rgb]{0.000, 0.690, 0.314}{\textbf{0.011}}} &0.016& \multicolumn{1}{c}{0.032} & \multicolumn{1}{c}{0.020} & 0.122  & \multicolumn{1}{c|}{0.062}   \\
		
			\hline
			& maxF  & \textcolor[rgb]{0.000, 0.690, 0.314}{\textbf{0.725}}&\textcolor[rgb]{1.000, 0.000, 0.000}{\textbf{0.791}}&\textcolor[rgb]{0.000, 0.439, 0.753}{\textbf{0.703}} & 0.697 & 0.620  & 0.655 & 0.585  & \multicolumn{1}{c}{0.603} & \multicolumn{1}{c}{0.640} & \multicolumn{1}{c}{0.614} & 0.608  & \multicolumn{1}{c}{0.572} & \multicolumn{1}{c}{0.520} & 0.464  & \multicolumn{1}{c|}{0.478}   \\
			DAVSOD\cite{fan2019shifting_ssav} & S-M   & \textcolor[rgb]{0.000, 0.690, 0.314}{\textbf{0.792}}&\textcolor[rgb]{1.000, 0.000, 0.000}{\textbf{0.846}}&0.770 & \textcolor[rgb]{0.000, 0.439, 0.753}{\textbf{0.779}} & 0.728  & 0.741 & 0.695  & \multicolumn{1}{c}{0.724} & \multicolumn{1}{c}{0.738} & \multicolumn{1}{c}{0.725} & 0.724  & \multicolumn{1}{c}{0.698} & \multicolumn{1}{c}{0.637} & 0.599  & \multicolumn{1}{c|}{0.624}   \\
			& MAE   & \textcolor[rgb]{0.000, 0.690, 0.314}{\textbf{0.064}} &\textcolor[rgb]{1.000, 0.000, 0.000}{\textbf{0.060}}&0.075& \textcolor[rgb]{0.000, 0.439, 0.753}{\textbf{0.070}} & 0.103  & 0.086 & 0.106  & \multicolumn{1}{c}{0.092 } & \multicolumn{1}{c}{0.084} & \multicolumn{1}{c}{0.096} & 0.092  & \multicolumn{1}{c}{0.116} & \multicolumn{1}{c}{0.159} & 0.220  & \multicolumn{1}{c|}{0.132}  \\
			
			\hline
			& maxF  & \textcolor[rgb]{1.000, 0.000, 0.000}{\textbf{0.822}}&0.660& \textcolor[rgb]{0.000, 0.439, 0.753}{\textbf{0.768}}& \textcolor[rgb]{0.000, 0.690, 0.314}{\textbf{0.781}} & 0.748 & 0.747  & 0.649  & \multicolumn{1}{c}{0.742 } & \multicolumn{1}{c}{0.735} & \multicolumn{1}{c}{0.724} & 0.735  & \multicolumn{1}{c}{0.742} & \multicolumn{1}{c}{0.670} & 0.690  & \multicolumn{1}{c|}{0.546}  \\
			VOS\cite{li2017benchmark} & S-M   & \textcolor[rgb]{0.000, 0.690, 0.314}{\textbf{0.844}}&0.741&\textcolor[rgb]{0.000, 0.439, 0.753}{\textbf{0.828}} & \textcolor[rgb]{1.000, 0.000, 0.000}{\textbf{0.845}} & 0.815  & 0.827 & 0.695  & \multicolumn{1}{c}{0.819} & \multicolumn{1}{c}{0.792} & \multicolumn{1}{c}{0.798} & 0.818  & \multicolumn{1}{c}{0.818} & \multicolumn{1}{c}{0.742} & 0.712  & \multicolumn{1}{c|}{0.624}  \\
			& MAE   & \textcolor[rgb]{0.000, 0.690, 0.314}{\textbf{0.060}}&0.074&0.069 & \textcolor[rgb]{1.000, 0.000, 0.000}{\textbf{0.052}} & 0.076  & \textcolor[rgb]{0.000, 0.439, 0.753}{\textbf{0.065}} & 0.115  & \multicolumn{1}{c}{0.073} & \multicolumn{1}{c}{0.075} & \multicolumn{1}{c}{\textcolor[rgb]{0.000, 0.439, 0.753}{\textbf{0.065}}} & 0.068 & \multicolumn{1}{c}{0.078} & \multicolumn{1}{c}{0.099} & 0.162  & \multicolumn{1}{c|}{0.145}   \\
			\hline
		\end{tabular}%
	}
	\label{table:comparison}%
\end{table*}%

Compared with the conventional ISOD, the patchwise SOD task here is much simpler because it has a relatively smaller problem domain.
Thus, we prefer off-the-shelf ISOD models with lightweight designs.
In our implementation, we choose the CPD~\cite{wu2019cascaded_cpd}, and we fine-tune it on the MSRA10K set, where we converted the original image-level training instances to patchwise instances.
The well-segmented patchwise saliency maps can be visualized in the second row of Fig.~\ref{fig:Online}.
Almost all salient objects or tiny parts have been well segmented by the patchwise refiner.
However, failure cases still exist, such as the middle case in the far left column, where only the dog's leg was detected, while the main body of the dog was missed.

\begin{figure}[t]
	\centering
	\includegraphics[width=1\linewidth]{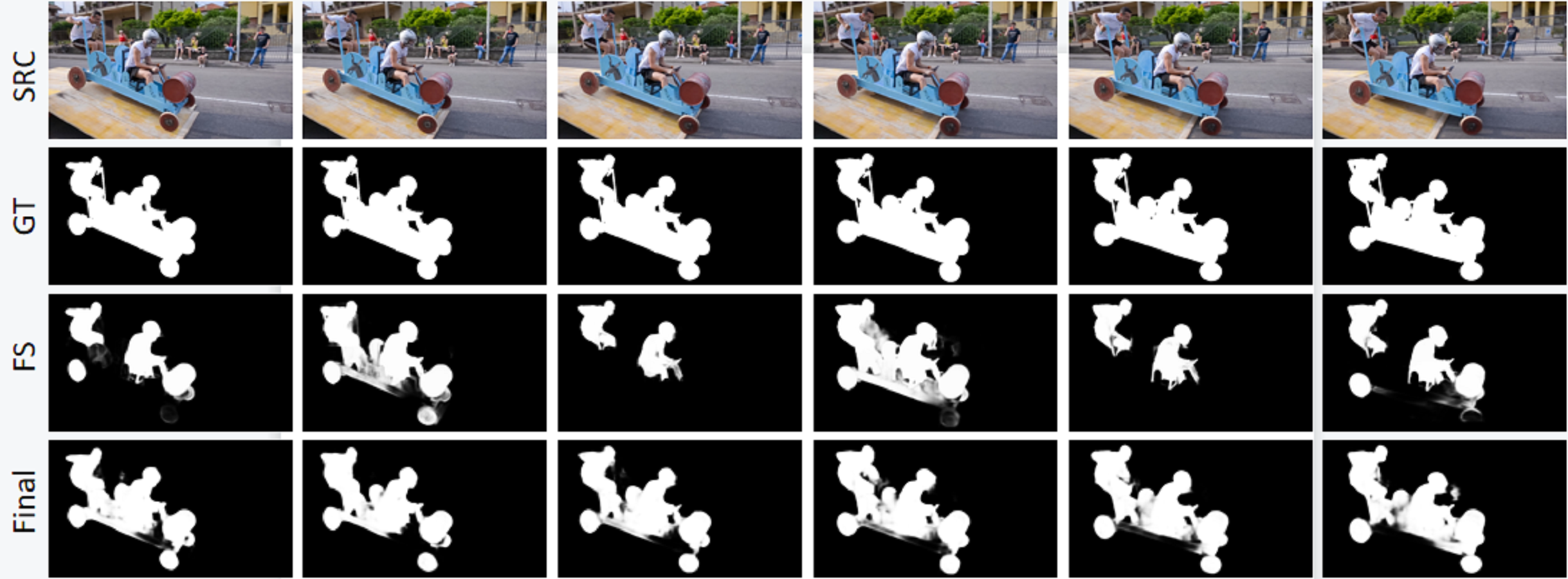}
    \vspace{-0.4cm}
	\caption{Qualitative comparison between the FS (Eq.~\ref{eq:negfil}) and the final results (`Final') obtained by the framewise refiner.}
\vspace{-0.4cm}
	\label{fig:OnlineQua}
\end{figure}

Intuitively, the frame-level saliency map can be easily obtained by `pasting' all its patchwise saliency maps.
However, due to the abovementioned failure cases, most of which are missing detections and false-alarm cases are very rare, the `pasting scheme' - the widely used average operation-based fusion - might not be suitable here, which can lead to frame-level saliency maps suffering from missing detections.
Therefore, we formulate the `pasting process' as follows:
\begin{equation}
FS_i = \max_j\Bigg\{proj\bigg(Z\Big(\mathcal{PR}\big(\Theta,crop(\textbf{I}_i,P_j)\big)\Big)\bigg)\Bigg\},
\label{eq:negfil}
\end{equation}
where $FS_i$ represents the frame-level saliency map of the $i$-th frame $\textbf{I}_i$; assume $\textbf{I}_i$ contains $b$ salient object proposals, the function $crop$ crops a patch from $\textbf{I}_i$ according to the coordinates provided by the object proposal $P_j$, and $1\leq j \leq b$; $\mathcal{PR}$ denotes the proposed patchwise refiner, $\Theta$ is its learnable hidden parameters, $\mathcal{PR}$ outputs the patchwise saliency map, $Z(\cdot)$ is a typical min-max normalization function, and function $proj$ pastes its input to the current video frame; thus, the output of this function is a matrix with the same height and width as the current frame.

The qualitative demonstration of $FS_i$ can be seen in the middle part of Fig.~\ref{fig:Online}.
Compared with the existing ISOD models~\cite{hou2019deeply_dss,CC-WZY-PR-RGBSal,qin2019basnet,CC-WZY-TMM-RGBSal,chen2020RGBDw,CC-RGBD-21-quality,CC-WGT-CVPR21,CC-MGX-RGBSal}, which consider only short-term information, our framewise saliency map is clearly built from the long-term perspective, which has a clear advantage in terms of robustness.

\subsection{Framewise Refinement via Online Model Fine-tuning}
\label{sec:FRVOMF}
Despite the merits mentioned before, framewise saliency maps ($FS$) still have two problems.
First, the quality of $FS$ is heavily dependent on the previous salient object proposal localization steps.
Second, this quality is also influenced by the `patchwise refiner'.
As a result, imperfect $FS$ can be observed frequently, such as the far-right column of Fig.~\ref{fig:Online}.
Thus, we propose the `framewise refiner', and only the most trustworthy $FS$ serves as the teacher to guide the online learning process.

\begin{table*}[t]
	\centering
	\caption{Component evaluation results. `SOPM': \textbf{s}alient \textbf{o}bject \textbf{p}roposal \textbf{m}ining scheme (Sec.~\ref{sec:SOPL}); `BC': \textbf{b}inary \textbf{c}lassifier (Sec.~\ref{sec:FLSOP}); `OLF': \textbf{o}n\textbf{l}ine \textbf{f}ine-tuning scheme (Sec.~\ref{sec:FRVOFT}); `MAX' and `AVE' respectively denote the maximizing operation based and the averaging based pasting process (Eq.~\ref{eq:negfil}). The corresponding qualitative demonstrations can be found in Fig.~\ref{fig:ComDemo}.}
\vspace{-0.2cm}
	\renewcommand\arraystretch{1.2}
	\resizebox{1\textwidth}{!}{
		\begin{tabular}{|l||c||c||c|c|c|c|c|c|c|c|c|c|c|c|c|}
			\hline
			\multicolumn{1}{|r||}{Dataset} & \multicolumn{3}{c|}{Davis~\cite{perazzi2016benchmark}} & \multicolumn{3}{c|}{Segv2~\cite{li2013video}} & \multicolumn{3}{c|}{Visal~\cite{wang2015consistent}} & \multicolumn{3}{c|}{DAVSOD~\cite{fan2019shifting_ssav}} & \multicolumn{3}{c|}{VOS~\cite{li2017benchmark}} \\
			
			\hline
			\multicolumn{1}{|r||}{Metrics} & \multicolumn{1}{c}{maxF} & \multicolumn{1}{c}{S-M} & MAE   & \multicolumn{1}{c}{maxF} & \multicolumn{1}{c}{S-M} & MAE   & \multicolumn{1}{c}{maxF} & \multicolumn{1}{c}{S-M} & MAE   & \multicolumn{1}{c}{maxF} & \multicolumn{1}{c}{S-M} & MAE   & \multicolumn{1}{c}{maxF} & \multicolumn{1}{c}{S-M} & MAE \\
			\hline
			
			\hline
			Motion Saliency (Eq.~\ref{eq:MSCompute})    & \multicolumn{1}{c}{0.798} & \multicolumn{1}{c}{0.854} & 0.044  & \multicolumn{1}{c}{0.648} & \multicolumn{1}{c}{0.760} & 0.054  & \multicolumn{1}{c}{0.627} & \multicolumn{1}{c}{0.738} & 0.079  & \multicolumn{1}{c}{0.450} & \multicolumn{1}{c}{0.613} & 0.148  & \multicolumn{1}{c}{0.405} & \multicolumn{1}{c}{0.566} & 0.167  \\
			CPD Baseline~\cite{wu2019cascaded_cpd}  & \multicolumn{1}{c}{0.778} & \multicolumn{1}{c}{0.859} & 0.032  & \multicolumn{1}{c}{0.778} & \multicolumn{1}{c}{0.841} & 0.023  & \multicolumn{1}{c}{0.941} & \multicolumn{1}{c}{0.942} & 0.016  & \multicolumn{1}{c}{0.608} & \multicolumn{1}{c}{0.724} & 0.092  & \multicolumn{1}{c}{0.735} & \multicolumn{1}{c}{0.818} & 0.068  \\
			+SOPM(MAX) & \multicolumn{1}{c}{0.878} & \multicolumn{1}{c}{0.891} & 0.025  & \multicolumn{1}{c}{0.852} & \multicolumn{1}{c}{0.882} & 0.021  & \multicolumn{1}{c}{0.930} & \multicolumn{1}{c}{0.924} & 0.019  & \multicolumn{1}{c}{0.684} & \multicolumn{1}{c}{0.757} & 0.076  & \multicolumn{1}{c}{0.789} & \multicolumn{1}{c}{0.827} & 0.063  \\
			+SOPM(MAX)+BC & \multicolumn{1}{c}{0.888} & \multicolumn{1}{c}{0.903} & 0.022  & \multicolumn{1}{c}{0.864} & \multicolumn{1}{c}{0.899} & 0.018  & \multicolumn{1}{c}{0.943} & \multicolumn{1}{c}{0.940} & 0.017  & \multicolumn{1}{c}{0.696} & \multicolumn{1}{c}{0.772} & 0.071  & \multicolumn{1}{c}{0.809} & \multicolumn{1}{c}{0.842} & 0.062  \\
			+SOPM(AVE)\ +BC & \multicolumn{1}{c}{0.883} & \multicolumn{1}{c}{0.899} & 0.023  & \multicolumn{1}{c}{0.862} & \multicolumn{1}{c}{0.895} & 0.019  & \multicolumn{1}{c}{0.943} & \multicolumn{1}{c}{0.939} & 0.017  & \multicolumn{1}{c}{0.685} & \multicolumn{1}{c}{0.764} & 0.073  & \multicolumn{1}{c}{0.808} & \multicolumn{1}{c}{0.841} & 0.062  \\
			+SOPM(MAX)+BC+OLF & \multicolumn{1}{c}{0.903} & \multicolumn{1}{c}{0.915} & 0.019  & \multicolumn{1}{c}{0.865} & \multicolumn{1}{c}{0.901} & 0.017 & \multicolumn{1}{c}{0.946} & \multicolumn{1}{c}{\textbf{0.947}} & 0.016  & \multicolumn{1}{c}{0.697} & \multicolumn{1}{c}{0.775} & 0.070  & \multicolumn{1}{c}{0.810} & \multicolumn{1}{c}{0.842} & \textbf{0.060}  \\
		+SOPM(MAX)+BC+OLF(KFS) & \multicolumn{1}{c}{\textbf{0.911}} & \multicolumn{1}{c}{\textbf{0.922}} & \textbf{0.016} & \multicolumn{1}{c}{\textbf{0.899}} & \multicolumn{1}{c}{\textbf{0.921}} & \textbf{0.013} & \multicolumn{1}{c}{\textbf{0.953}} & \multicolumn{1}{c}{\textbf{0.947}} & \textbf{0.011} & \multicolumn{1}{c}{\textbf{0.725}} & \multicolumn{1}{c}{\textbf{0.792}} & \textbf{0.064} & \multicolumn{1}{c}{\textbf{0.822}} & \multicolumn{1}{c}{\textbf{0.844}} & \textbf{0.060} \\
			\hline
		\end{tabular}%
	}
	\label{table:component}%
\end{table*}%

In practice, the framewise refiner can be any ISOD model using only spatial information, where we simply continue using the CPD as the framewise refiner due to its lightweight design.
Note that if the framewise refiner is taught by some of the high-quality $FS$, it can perform very well on the remaining unseen frames, producing very high-quality saliency maps.
We show the difference between $FS$ and the final saliency map in Fig.~\ref{fig:OnlineQua}.

Outwardly, using only the high-quality $FS$ to fine-tune the framewise refiner omits the temporal information completely.
Nevertheless, it should be noted that the temporal information is implicitly embedded in the $FS$ because the previous salient object proposal localization steps fully consider the motion saliency cues (i.e., Eq.~\ref{eq:PMS}).

Let us now move to the details regarding how to select the high-quality $FS$ --- the proposed fast keyframe selection scheme.
Compared with spatial cues, motion cues are more likely to attract our visual attention.
Thus, in some cases, motion saliency cues can be a very effective indicator for locating real salient objects.
However, the motion saliency cue has one critical drawback, i.e., it is unstable in essence, and we noted this issue multiple times before.
Therefore, we consider both spatial and temporal saliency cues to derive robust VSOD.

Considering the common attribute of high-quality $FS$, both their corresponding spatial and temporal saliency maps tend to be high-quality maps, and the reverse usually holds.
Hence, for each frame, we can use the similarity between the spatial saliency ($FS$) and temporal saliency ($MS$) to measure the quality degree of $FS$ because those frames with low-quality $FS$ are unlikely to have an `extremely' high spatiotemporal saliency consistency.

To be more specific, our keyframe selection process can be summarized into the following two steps:\\
First, for each frame, we compute the consistency between its $FS$ and $MS$, where we use the S-measure~\cite{fan2017structure} as the similarity measurement because $MS$ focuses more on the overall structural similarity rather than the tiny saliency details that are quite rare in $MS$.\\
Second, for every $B$ frames (the exact choice of $B$ can be found in the ablation study), only the one with the largest spatiotemporal saliency consistency degree is selected as the key frame.
For example, given a sequence with $N$ frames in total, there are $N/B$ keyframes to be selected, and all these keyfames are used to fine-tune the framewise refiner.
The fine-tuning process can be performed very quickly (10 epochs), and the final VSOD results can be obtained by taking each nonkeyframe as the input of the framewise refiner.
To maintain strong generalization ability, for each sequence, the framewise refiner is recovered to its initial status.

To facilitate readers' understanding, we have provided a method flow chat, where all key steps have been included into it.
The detailed flow chat can be found in \textbf{Algorithm 1}.

\begin{algorithm}[!h]
	\caption{Key Steps of Our Approach.}
	\renewcommand{\algorithmicrequire}{\textbf{Input:}}
	\renewcommand{\algorithmicensure}{\textbf{Output:}}
	\begin{algorithmic}[1]
		\REQUIRE a video sequence;
        \ENSURE high-quality VSOD results;
        \STATE object proposal computation via the off-the-shelf tool SEOD;
        \STATE color saliency computation using the off-the-shelf tool CPD;
        \STATE optical flow computation (Eq. 5);
        \STATE \textbf{\emph{train}} motion saliency model (Eq. 6) and output motion saliency maps (Eq. 7);
        \STATE cluster all object proposals into $K$ clusters (Eq. 1);
        \STATE coarsely localize $v$ salient clusters (Eq.~2-4); \\
               \textbf{For} \emph{iter}=1:3
        \STATE rank each cluster's object proposals (Eq. 11-12) via SIM (Eq. 10) and PMS (Eq. 8);
        \STATE dynamically threshold each cluster's object proposals (Eq. 13-16);
        \STATE formulate pos. and neg. training samples (Eq. 17-18);
        \STATE \textbf{\emph{train}} binary classifier (Eq. 9);\\
               \textbf{End}
        \STATE \textbf{\emph{train}} patch refiner using Pos. (identical to Eq. 6);
        \STATE generate frame-level saliency map via patch refiner (Eq. 19);
        \STATE select a key frame from each $B$ frames (Sec. IV-B);
        \STATE \textbf{\emph{train}} frame-wise refiner using the selected key frames (identical to Eq. 6);
        \STATE utilize frame-wise refiner to output VSOD result;
	\end{algorithmic}
	\label{alg:label1}
\end{algorithm}

\begin{figure*}[t]
	\centering
	\includegraphics[width=1\linewidth]{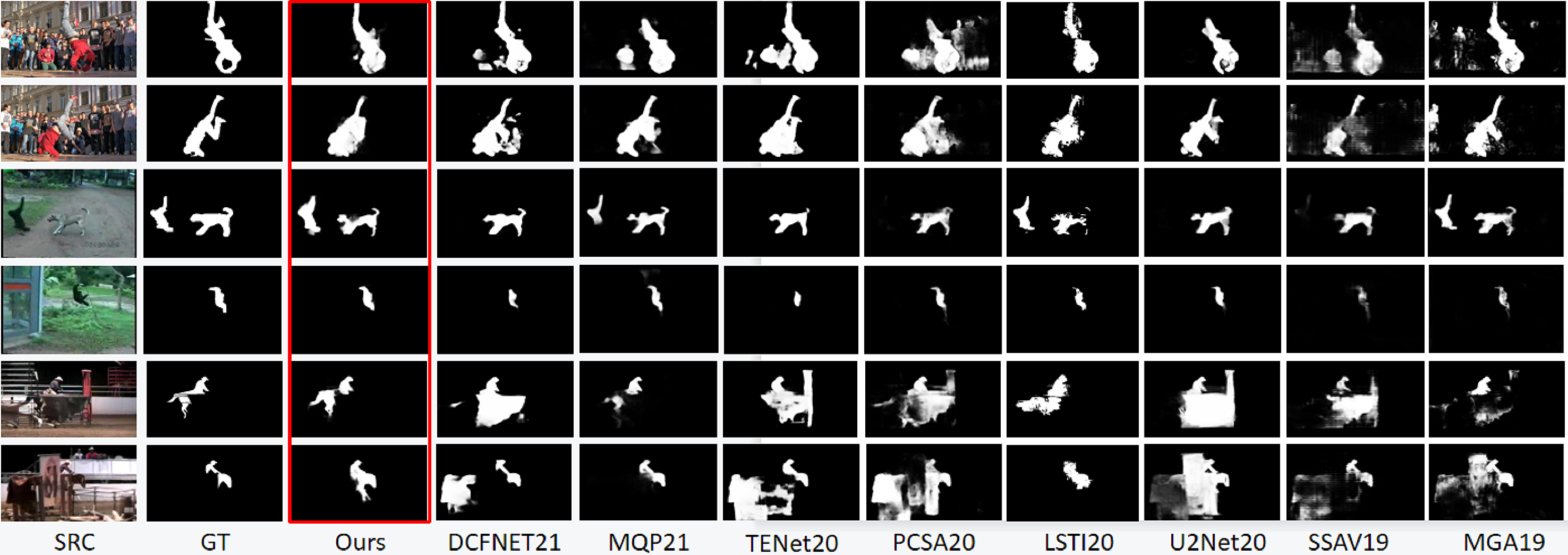}
\vspace{-0.3cm}
	\caption{Qualitative comparisons with the current SOTA methods. Due to the limited space, we only list several of the most representative comparisons here.\vspace{-0.6cm}}
	\label{fig:OverallCompare}
\end{figure*}
%\textbf{Acknowledgments}.This work was supported by National Natural Science Foundation of China (Grant No. 61802215 and 61806106) and Natural Science Foundation of Shandong Province (No. ZR2019BF011 and ZR2019QF009).
\section{Experiments}
\subsection{Datasets}
Following the conventional experimental setting, we evaluated the proposed approach on five widely used publicly available datasets, including Davis~\cite{perazzi2016benchmark}, Segtrack-v2~\cite{li2013video}, VISal~\cite{wang2015consistent}, DAVSOD~\cite{fan2019shifting_ssav}, and VOS~\cite{li2017benchmark}.
\begin{itemize}
	\item
	The Davis dataset contains 50 video sequences with 3,455 frames in total, and most of its sequences only contain moderate motions.
	\item
	The Segtrack-v2 dataset contains 13 video sequences (excluding the penguin sequence) with 1,024 frames in total, containing complex backgrounds and variable motion patterns, which is generally more challenging than the Davis dataset. This dataset is dominated by a temporal source with fast object movements.
	\item
	The VISal dataset contains 17 video sequences with 963 frames in total, and this dataset is relatively simple. This dataset is dominated by spatial sources, where most of the existing SOTA ISOD models using spatial information alone performed very well on this dataset.
	\item
	The DAVSOD dataset contains 226 video sequences with 23,938 frames in total, which is the most challenging dataset, involving various object instances, different motion patterns, and saliency shifting between different objects. Since most SOTA VSOD models have failed to consider the attention shifting mechanism, their quantitative scores over this set are very low.
	\item
	The VOS dataset contains 40 video sequences with 24,177 frames in total, yet only 1,540 frames were annotated well, in which the sequences were all obtained in indoor scenes.
\end{itemize}

\subsection{Implementation Details}
We implemented our method on a PC with an Intel(R) Xeon(R) CPU, NVIDIA GTX2080Ti GPU (with 11G RAM) and 64G RAM.
We choose SEOD~\cite{tan2020efficientdet} as the object detector to obtain object proposals.
The patchwise and framewise refiners (Sec.~\ref{sec:FRVOFT} and Fig.~\ref{fig:Online}) follow an identical structure to the CPD~\cite{wu2019cascaded_cpd}.
The patchwise refiner is retrained on patchwise data generated from the widely used MSRA10K~\cite{ChengPAMI}.
The framewise refiner follows the online learning scheme, which is trained on DAVIS-TR~\cite{perazzi2016benchmark} in advance, and then fine-tuned online on the keyframes mined from the current input sequence.
Note that for each input video sequence, we perform the online fine-tuning process only once.
The motion saliency network (Eq.~\ref{eq:ISODFinetune}) also follows an identical structure to the CPD, which is initially trained on the MSRA10K set and then fine-tuned on the DAVIS-TR set.
The binary classifier is a multilayer fully connected network whose network structure follows the lightweight ResNet18, and the updating process takes 20 epochs.
To avoid overfitting problems, we adopted random horizontal flips for data augmentation.

\subsection{Evaluation Metrics}
To accurately measure the consistency between the predicted VSOD and the manually annotated ground truth, we adopt five commonly used evaluation metrics, including the precision rate, recall rate, maximum F-measure value (maxF)~\cite{achanta2009frequency}, mean absolute error (MAE)~\cite{perazzi2012saliency}, and structure measure value (S-measure)~\cite{fan2017structure}.

\begin{figure*}[t]
	\centering
	\includegraphics[width=\linewidth]{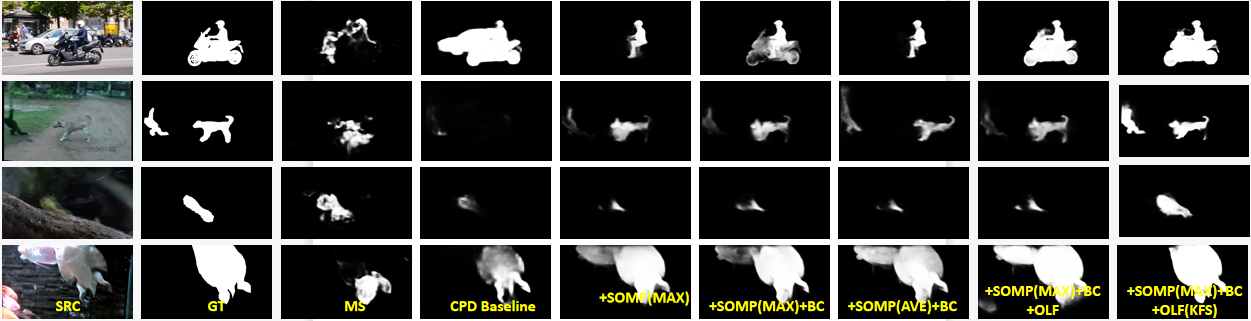}
\vspace{-0.6cm}
	\caption{Qualitative demonstrations regarding different major component. The corresponding quantitative results of this demonstration can be found in Table~\ref{table:component}.\vspace{-0.4cm}}
	\label{fig:ComDemo}
\end{figure*}

\begin{table}[http]
	\centering
	\caption{More quantitative evidence regarding the effectiveness of the proposed iterative mining scheme. `Iter0-PA' and `Iter0-OL' respectively denote the performance of $FS$ (Eq.~\ref{eq:negfil}) and the final result (Sec.~\ref{sec:FRVOMF}) in the `Iteration 0'.}
\vspace{-0.2cm}
\Large
	\renewcommand\arraystretch{1.2}
	\resizebox{1\linewidth}{!}{
		\begin{tabular}{|cc|cc|cc|cc|cc|}
			\hline
			\multicolumn{1}{|c}{Sets} & Metrics & Iter0-PA & Iter0-OL & Iter1-PA & Iter1-OL & Iter2-PA & Iter2-OL & Iter3-PA & Iter3-OL \\
			\hline
			\hline
			\multirow{3}[1]{*}{\begin{sideways}Davis\ \ \end{sideways}}
			& maxF  & 0.878  & 0.892  & 0.902  & 0.908  & 0.899  & 0.884  & \textbf{0.907}  &          \textbf{0.911}  \\
		    & S-M   & 0.907  & 0.909  & 0.910  & 0.921  & 0.909  & 0.921  & \textbf{0.916}  & \textbf{0.922}  \\
		    & MAE   & 0.017  & 0.020  & 0.017  & 0.017  & 0.018  & \textbf{0.016}  & \textbf{0.016}  & \textbf{0.016}  \\
			\hline
			\multirow{3}[1]{*}{\begin{sideways}Segv2\ \ \end{sideways}}
			& maxF  & 0.843  & 0.855  & 0.848  & 0.874  & 0.848  & 0.866  & \textbf{0.849}  & \textbf{0.899}  \\
		    & S-M   & 0.881  & 0.894  & 0.888  & 0.905  & 0.889  & 0.900  & \textbf{0.890}  & \textbf{0.921}  \\
			& MAE   & 0.021  & 0.018  & 0.020  & 0.016  & \textbf{0.019}  & 0.016  & \textbf{0.019}  & \textbf{0.013}  \\
			\hline
			\multirow{3}[1]{*}{\begin{sideways}VISal\ \ \end{sideways}}
			& maxF  & 0.927  & 0.931  & \textbf{0.945}  & 0.952  & 0.944  & 0.948  & 0.943  & \textbf{0.953}  \\
			& S-M   & 0.921  & 0.930  & 0.941  & 0.943  & 0.940  & 0.945  & \textbf{0.942}  & \textbf{0.947}  \\
		    & MAE   & 0.018  & 0.014  & 0.016  & 0.012  & 0.016  & 0.012  & \textbf{0.015}  & \textbf{0.011}  \\
			\hline
			\multirow{3}[1]{*}{\begin{sideways}{ DAVSOD \ \ }\end{sideways}}
			 & maxF  & 0.669  & 0.646  & 0.694  & 0.679  & 0.696  & 0.675  & \textbf{0.702}  & \textbf{0.725}  \\
			 & S-M   & 0.739  & 0.734  & 0.768 & 0.763  & 0.769  & 0.763  & \textbf{0.775}  & \textbf{0.792}  \\
			 & MAE   & 0.072 & 0.077 & 0.071  & 0.070  & \textbf{0.070}  & 0.071  & 0.071  & \textbf{0.064}  \\
			\hline
			\multirow{3}[1]{*}{\begin{sideways}VOS\ \ \end{sideways}}
			 & maxF  & 0.817  & 0.785  & \textbf{0.830}  & 0.805  & 0.828  & 0.814  & 0.821  & \textbf{0.822}  \\
			 & S-M   & 0.845  & 0.827  & \textbf{0.857}  & 0.841  & 0.855  & \textbf{0.849}  & 0.850  & 0.844  \\
			 & MAE   & 0.050  & 0.063  & \textbf{0.049}  & 0.057  & 0.051  & \textbf{0.053}  & 0.058  & 0.060  \\
			\hline
		\end{tabular}%
	}
	\label{table:overallpertwo}%
\end{table}%

\begin{table*}[!b]
	\centering
\vspace{-0.2cm}
	\caption{Ablation study on the online fine-tuning epochs for the `Framewise Refiner' (Sec.~\ref{sec:FRVOMF}).}
\vspace{-0.2cm}
	\renewcommand\arraystretch{1.2}
	\resizebox{0.9\textwidth}{!}{
		\begin{tabular}{|c|ccccccccc|}
			\hline
			Metrics & 2-epoches & 4-epoches & 6-epoches & 8-epoches & 10-epoches & 12-epoches & 14-epoches & 16-epoches & 18-epoches \\
			\hline
			\hline
			maxF  & 0.795 & 0.823  & 0.864  & 0.883  & 0.886  & 0.895  & \textbf{0.899}  & 0.894  & 0.894  \\
			S-M   & 0.855  & 0.874  & 0.902  & 0.912  & 0.913  & 0.918  & \textbf{0.921}  & 0.919  & 0.919  \\
			MAE   & 0.034  & 0.029  & 0.015  & 0.014  & 0.014  & \textbf{0.013}  & \textbf{0.013}  & \textbf{0.013}  & \textbf{0.013}  \\
\hline
			Times (Seconds) & 2.000  & 4.000  & 6.000  & 8.000  & 10.00  & 12.00  & 14.00  & 16.00  & 18.00  \\
			\hline
		\end{tabular}%
	}
	\label{tabel:epoch}%
\end{table*}%

\subsection{Quantitative Comparisons}
We compared our method with 14 SOTA approaches (see Table~\ref{table:comparison}),
including  DCFNet~\cite{zhang2021dynamic}, MQP~\cite{chen2021novel}, TENet20~\cite{ren20TENet},  U2Net20~\cite{qin2020u2}, PCSA20~\cite{gupyramid_pcsa}, LSTI20~\cite{chen2019improved}, SSAV19~\cite{fan2019shifting_ssav}, MGA19~\cite{li2019motion_mga}, COS19~\cite{lu2019see}, CPD19~\cite{wu2019cascaded_cpd}, PDB18~\cite{song2018pyramid_pdbm}, MBN18~\cite{li2018unsupervised}, SCO18~\cite{chen2018scom}, and SFLR17~\cite{chen17ST}.
Among all the SOTA competitors compared here, TENet20 and LSTI20 are mainly designed for introducing more long-term information into the current problem domain.
However, compared with these two models, the long-term attribute of our approach is clearly stronger because all object proposals in the input sequences are simultaneously available to the current problem, while the basic methodology of both TENet20 and LSTI20 might be short-term in essence.
For example, the graph structure adopted by TENet20 is a clear short-term manner, and the long-term alignment process of LSTI20 can be quite limited because long-term spatial alignment might be very difficult if the spatial appearance of the salient object has changed significantly.

Compared with other short-term models (e.g., PCSA20, SSAV19, and PDB18), our method outperformed them significantly, especially on the SegTrack-v2 set, where the salient objects tend to exhibit fast object movements and vary their spatial appearances very quickly; thus, the conventional short-term methodology can easily produce failure detections if both the current spatial and temporal saliency cues are incorrect.
In sharp contrast, the proposed long-term method can to utilize the beyond-scope information to help the current saliency prediction.

Specifically, we noticed that our method failed to achieve the best performance on the VISal set.
This can be explained by the attribute of the VISal set --- all video sequences are dominated by spatial saliency cues, while the temporal saliency cues become completely helpless.
This issue can be evidenced by the fact that the ISOD model (i.e., the CPD19), which only considers the spatial information, could still perform very well on the VISal set. In addition, we have presented the qualitative results in Fig.~\ref{fig:OverallCompare}

\subsection{Component Evaluations}
In this subsection, we verify the effectiveness of each major component in the proposed approach, and the quantitative results are detailed in Table~\ref{table:component}.

Both baseline models, i.e., the motion saliency deep model (Eq.~\ref{eq:ISODFinetune}) and the original CPD~\cite{wu2019cascaded_cpd}, exhibit the worst results.
Then, by using the proposed clustering-based coarse localization (i.e., the `+SOPM(MAX)') to find salient object proposals, the overall performance, which is already comparable to some SOTA models published in 2019, can be improved significantly.
Then, by using the proposed data mining scheme, i.e., the binary classifier, the overall performance can be further improved.
Quantitative evidence can be seen by comparing `+SOPM(MAX)' with `+SOPM(MAX)+CL', showing the effectiveness of the proposed iterative mining scheme.
Since the binary classifier-based data mining scheme is the major component of this paper, we provide an additional quantitative evaluation to further verify its effectiveness in the next subsection.

In our implementation, all object proposal-based saliency maps are pasted back to the original video frame via the maximizing operation (Eq.~\ref{eq:negfil}), and we also test the averaging operation, and the result is shown as `+SOPM(AVE)+CL'.
Clearly, the maximizing operation can more suit the pasting scheme than the averaging operation.

To verify the effectiveness of the proposed online fine-tuning scheme, we test the `+SOPM(MAX)+CL+OLF', where, for each five video frames, one frame is randomly selected as the key frames, and the `framewise refiner' will be fine-tuned on these key frames using their $FS$ (Eq.~\ref{fig:Online}) as the pseudoGTs.
Because this random selection process easily introduces some less trustworthy $FS$ into the fine-tuning process, the performance gain achieved by this online updating is very marginal.
By using the proposed keyframe selection strategy, which has been denoted as `+SOPM(MAX)+CL+OLF(KFS)', the overall performance can be improved by an average of 2\%.

\subsection{More Quantitative Evidence of the Effectiveness of the Proposed Iterative Mining Scheme}
%We test the overall performance of the binary classifier in each mining iteration.
%As seen in Table~\ref{table:clsper}, we demonstrate the precision, recall, and F-measure rate on two groups of object proposals, where all object proposals are divided into either the salient group or the nonsalient group according to the pixelwise GT (i.e., an object proposal is regarded as salient if more than 20\% of the pixels inside it are salient).
%The `Iteration 0' represents the initial localization via Eq.~\ref{eq:posfil} and Eq.~\ref{eq:negfil}.
%`Iterations 1-3' show the overall performance of the binary classifier-based iterative mining scheme (Sec.~\ref{sec:IMMTOP}).
%Clearly, the overall performance improves steadily with the continuation of mining iterations.

The overall performances of the final saliency maps obtained by using or not using the online fine-tuning scheme are shown in Table~\ref{table:overallpertwo}, where `Iter0-PA' and `Iter0-OL' denote the performance of $FS$ (Eq.~\ref{eq:negfil}) and the final result (Sec.~\ref{sec:FRVOMF}) in the `Iteration 0', respectively.
Specifically, we also observed that the overall performance seems to reach a plateau after `Iteration 3', where some quantity metrics decrease slightly, e.g., the maxF on the Segv2 set (0.893$\rightarrow$0.885).
Therefore, we decided to omit `Iteration 4' and set the maximum iteration number to 3 to ensure computational efficiency.

\subsection{Ablation Study on the Online Fine-tuning Epochs}
Actually, the proposed network online fine-tuning process (Sec.~\ref{sec:FRVOMF}) costs some additional computational time.
However, in our case, we adapt only the framewise refiner to the current video sequence, where the training data are relatively small.
In addition, to avoid overfitting and consider the fact that the framewise refiner performs very well if it has weakly learned by taking the $FS$ (Eq.~\ref{eq:negfil}) of those trustworthy frames as the learning objective, the online fine-tuning process only requires several epochs.
As shown in Table~\ref{tabel:epoch}, we chose 10 epochs as the optimal choice to ensure the overall computational efficiency because the performance gains achieved by using some additional epochs could be very marginal.

\subsection{Ablation Study on the Clustering Parameter $K$}
The hyperparameter $K$ determines the initial cluttering number, and we have tested multiple choices, \emph{i.e.}, $K$=\{4,6,8,10\}, where the $K$=8 is the empirically selected choice in the previous version.
The quantitative results have been shown in Table~\ref{abK}.
Clearly, $K$=8 is the best choice, where other choices of $K$ may lead to some performance degenerations.
Specifically, we have noticed that overall performance decreased significantly when $K$ = 10 over the SegTrack set.
The main reason is that the entire \emph{birdfall} sequence is completely ill-detected.
Since the falling bird in this low-resolution sequence is a quite small size object, our clustering process has assigned multiple background proposals with similar appearances to the bird into the salient clusters.
\emph{W.r.t.} other choices of $K$, the overall performances tend to stay stable.
Also, another reason for the $K$=8 to be the best choice might be the fact that all other parameters (\emph{e.g.}, the mining iteration times) are selected under the default $K$=8. Thus, it is quite reasonable for other choices of $K$ to exhibit inferior results.

\begin{table}[!t]
	\centering
	\caption{Ablation study on the pre-given clustering number $K$, where we have tested $K$=\{4,6,8,10\}, and we set the $K$=8 as the optimal choice.}
\vspace{-0.2cm}
	\renewcommand\arraystretch{1.2}
	\resizebox{1\linewidth}{!}{
		\begin{tabular}{|l||c||c||c|c|c|c|c|c|c|}
			\hline
			\multicolumn{1}{|r||}{Dataset} & \multicolumn{3}{c|}{Davis~\cite{perazzi2016benchmark}} & \multicolumn{3}{c|}{Segv2~\cite{li2013video}} & \multicolumn{3}{c|}{Visal~\cite{wang2015consistent}}\\
			
			\hline
			\multicolumn{1}{|r||}{Metrics} & \multicolumn{1}{c}{maxF} & \multicolumn{1}{c}{S-M} & MAE   & \multicolumn{1}{c}{maxF} & \multicolumn{1}{c}{S-M} & MAE   & \multicolumn{1}{c}{maxF} & \multicolumn{1}{c}{S-M} & MAE\\
			\hline
			
			\hline
$K$=4 & \multicolumn{1}{c}{0.893} & \multicolumn{1}{c}{0.908} & 0.019  & \multicolumn{1}{c}{0.890} & \multicolumn{1}{c}{0.912} & 0.014  & \multicolumn{1}{c}{0.951} & \multicolumn{1}{c}{0.947} & 0.013 \\
$K$=6 & \multicolumn{1}{c}{0.905} & \multicolumn{1}{c}{0.915} & 0.017  & \multicolumn{1}{c}{0.894} & \multicolumn{1}{c}{0.917} & 0.014  & \multicolumn{1}{c}{0.949} & \multicolumn{1}{c}{0.947} & 0.013 \\
\cellcolor[rgb]{.92, .92, .92}$K$=8 & \multicolumn{1}{c}{{0.911}} & \multicolumn{1}{c}{{0.922}} & {0.016} & \multicolumn{1}{c}{{0.899}} & \multicolumn{1}{c}{{0.921}} & {0.013} & \multicolumn{1}{c}{{0.953}} & \multicolumn{1}{c}{{0.947}} & {0.011} \\
$K$=10 & \multicolumn{1}{c}{{0.912}} & \multicolumn{1}{c}{{0.923}} & {0.015} & \multicolumn{1}{c}{{0.853}} & \multicolumn{1}{c}{{0.891}} & {0.021} & \multicolumn{1}{c}{{0.952}} & \multicolumn{1}{c}{{0.949}} & {0.011} \\
			\hline
		\end{tabular}%
	}
	\label{abK}%
\end{table}%

\subsection{Ablation Study on the Key Frame Sampling Parameter $B$}
The hyper parameter $B$ is the frame batch size, which determines the overall keyframe number to be used during the model fine-tuning process.
Generally, a large choice of $B$ correlates to sparse key frame selections, and a small choice of $B$ will lead to a dense sampling of keyframes.
The exact ablation study towards $B$ can be seen in Table~\ref{abB}, where we have tested multiple choices, including $B$=\{3,5,8,10\}, and $B$=5 is the default choice.
As shown in the table, the overall performances with different choices of $B$ are very stable, even though $B$=5 slightly outperforms other choices.
The main reason for this phenomenon is that there exists a clear trade-off, \emph{i.e.}, a large number of keyframes may lead the fine-tuning process difficult to reach convergency, while a small number of keyframes may result in over-fitting.
Therefore, we continue to use $B$=5 as the best choice.

\begin{table}[!t]
	\centering
	\caption{Ablation study on the key frame sampling parameter $B$, where we have tested $B$=\{3,5,8,10\}, and we set the $B$=5 as the optimal choice, \emph{i.e.}, there will be at least 1 frame for each 5 frames will be selected as the default key frames during the online model refining process.}
\vspace{-0.2cm}
	\renewcommand\arraystretch{1.2}
	\resizebox{1\linewidth}{!}{
		\begin{tabular}{|l||c||c||c|c|c|c|c|c|c|}
			\hline
			\multicolumn{1}{|r||}{Dataset} & \multicolumn{3}{c|}{Davis~\cite{perazzi2016benchmark}} & \multicolumn{3}{c|}{Segv2~\cite{li2013video}} & \multicolumn{3}{c|}{Visal~\cite{wang2015consistent}}\\
			
			\hline
			\multicolumn{1}{|r||}{Metrics} & \multicolumn{1}{c}{maxF} & \multicolumn{1}{c}{S-M} & MAE   & \multicolumn{1}{c}{maxF} & \multicolumn{1}{c}{S-M} & MAE   & \multicolumn{1}{c}{maxF} & \multicolumn{1}{c}{S-M} & MAE\\
			\hline
			\hline
$B$=3 & \multicolumn{1}{c}{0.909} & \multicolumn{1}{c}{0.918} & 0.016  & \multicolumn{1}{c}{0.885} & \multicolumn{1}{c}{0.909} & 0.014  & \multicolumn{1}{c}{0.952} & \multicolumn{1}{c}{0.947} & 0.011 \\
 \cellcolor[rgb]{.92, .92, .92}$B$=5 & \multicolumn{1}{c}{{0.911}} & \multicolumn{1}{c}{{0.922}} & {0.016} & \multicolumn{1}{c}{{0.899}} & \multicolumn{1}{c}{{0.921}} & {0.013} & \multicolumn{1}{c}{{0.953}} & \multicolumn{1}{c}{{0.947}} & {0.011} \\
$B$=8 & \multicolumn{1}{c}{{0.908}} & \multicolumn{1}{c}{{0.918}} & {0.017} & \multicolumn{1}{c}{{0.842}} & \multicolumn{1}{c}{{0.892}} & {0.015} & \multicolumn{1}{c}{{0.951}} & \multicolumn{1}{c}{{0.948}} & {0.011} \\
$B$=10 & \multicolumn{1}{c}{{0.902}} & \multicolumn{1}{c}{{0.917}} & {0.018} & \multicolumn{1}{c}{{0.848}} & \multicolumn{1}{c}{{0.898}} & {0.016} & \multicolumn{1}{c}{{0.949}} & \multicolumn{1}{c}{{0.948}} & {0.013} \\
			\hline
		\end{tabular}%
	}
\label{abB}
\end{table}%

\begin{table}[!t]
	\centering
	\caption{Quantitative results of our method after using other object proposal (YOLOv4~\cite{bochkovskiy2020yolov4}) and optical flow (LiteFlow~\cite{Hui_2018_CVPR}) methods.}
\vspace{-0.2cm}
	\renewcommand\arraystretch{1.2}
	\resizebox{1\linewidth}{!}{
		\begin{tabular}{|l||c||c||c|c|c|c|c|c|c|}
			\hline
			\multicolumn{1}{|r||}{Dataset} & \multicolumn{3}{c|}{Davis~\cite{perazzi2016benchmark}} & \multicolumn{3}{c|}{Segv2~\cite{li2013video}} & \multicolumn{3}{c|}{Visal~\cite{wang2015consistent}}\\
			
			\hline
			\multicolumn{1}{|r||}{Metrics} & \multicolumn{1}{c}{maxF} & \multicolumn{1}{c}{S-M} & MAE   & \multicolumn{1}{c}{maxF} & \multicolumn{1}{c}{S-M} & MAE   & \multicolumn{1}{c}{maxF} & \multicolumn{1}{c}{S-M} & MAE\\
			\hline
			
			\hline
MS Baseline & \multicolumn{1}{c}{0.798} & \multicolumn{1}{c}{0.854} & 0.044  & \multicolumn{1}{c}{0.648} & \multicolumn{1}{c}{0.760} & 0.054  & \multicolumn{1}{c}{0.627} & \multicolumn{1}{c}{0.738} & 0.079 \\
			\hline
			SEOD~\cite{tan2020efficientdet}$\rightarrow$YOLOv4~\cite{bochkovskiy2020yolov4}    & \multicolumn{1}{c}{0.907} & \multicolumn{1}{c}{0.919} & 0.016  & \multicolumn{1}{c}{0.858} & \multicolumn{1}{c}{0.899} & 0.017  & \multicolumn{1}{c}{0.933} & \multicolumn{1}{c}{0.931} & 0.018\\
			FlowNet~\cite{sun2018pwc}$\rightarrow$LiteFlow~\cite{Hui_2018_CVPR}    & \multicolumn{1}{c}{0.910} & \multicolumn{1}{c}{0.922} & 0.015  & \multicolumn{1}{c}{0.887} & \multicolumn{1}{c}{0.913} & 0.014  & \multicolumn{1}{c}{0.941} & \multicolumn{1}{c}{0.939} & 0.014\\
			\hline
		SEOD~\cite{tan2020efficientdet}+FlowNet~\cite{sun2018pwc}& \multicolumn{1}{c}{{0.911}} & \multicolumn{1}{c}{{0.922}} & {0.016} & \multicolumn{1}{c}{{0.899}} & \multicolumn{1}{c}{{0.921}} & {0.013} & \multicolumn{1}{c}{{0.953}} & \multicolumn{1}{c}{{0.947}} & {0.011} \\
			\hline

		\end{tabular}%
	}
	\label{table:opticalflowandobjectproposal}%
\end{table}%

\begin{table}[!b]
	\centering
\vspace{-0.4cm}
	\caption{Detailed averaged time cost for a single frame. This result was obtained on a PC with an Intel(R) Xeon(R) CPU, NVIDIA GTX2080Ti GPU (with 11G RAM) and 64G RAM. This experiment was carried out on the Visal set only.}
\vspace{-0.2cm}
	\begin{tabular}{|l|c|}
		\hline
		MainSteps & Seconds \\
		\hline
		\hline
        Optical Flow (Sec. III-A) & 0.060s\\
        Object Proposal (Sec. III-A) & 0.121s\\
        Low-level Saliency (Sec. III-A) & 0.050s\\
        Clustering and Ranking (Sec. III-B) & 0.020s  \\
		Binary Classifier Training \& Updating (Sec. III-D) & 0.057s  \\
		Patchwise Refiner (Sec. IV-A) & 0.030s  \\
		Framewise Refiner (Sec. IV-B) & 0.080s  \\
        \hline
		Total & 0.418s \\
		\hline
	\end{tabular}%
	\label{tab:time}%
\vspace{-0.4cm}
\end{table}%

\subsection{How will Different Object Proposal and Optical Flow Methods Affect the Overall Performance}
Actually, both object proposal and optical flow methods affect the overall performance of our method, of course. And the main reason is clear, \emph{i.e.}, all other parameters are assigned based on the combination of SEOD + FlowNet, where new choices of object proposal and optical flow may lead our model to stay in the sub-optimal situation.
As shown in Table~\ref{table:opticalflowandobjectproposal}, using either YOLOv4 or LiteFlow could lead to some performance degenerations as expected. However, the quantitative results illustrated in Table~\ref{table:opticalflowandobjectproposal} also suggest that our approach is generally stable. Though the overall performance has been decreased, the overall performances of the two modified versions (\emph{i.e.}, YOLOv4+FlowNet and SEOD+LiteFlow) still significantly outperform the \textbf{m}otion \textbf{s}aliency (MS) baseline.
Also, this experiment has been included in the revised version.

\subsection{Failure Cases and Limitations}
\begin{figure}[!t]
	\centering
	\includegraphics[width=1\linewidth]{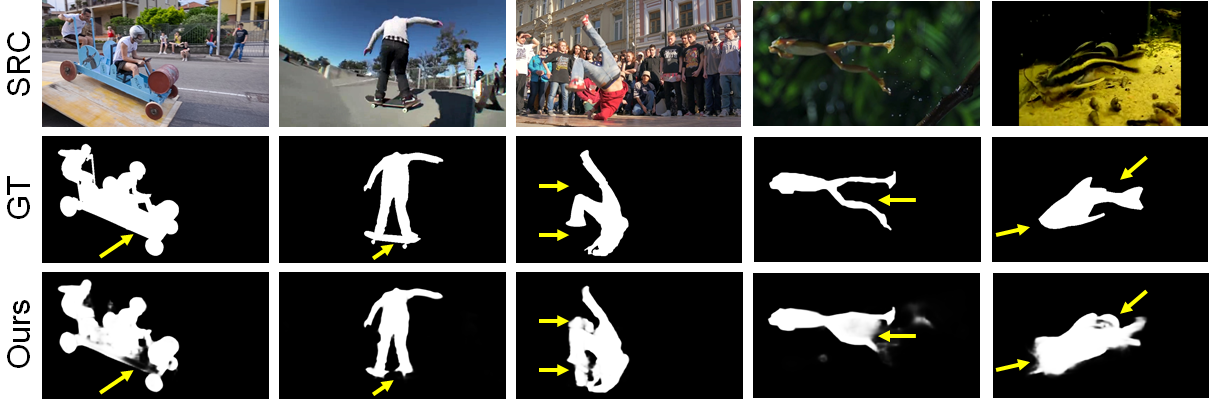}
    \vspace{-0.6cm}
	\caption{Demonstrations of some of the most representative failure cases.}
    \vspace{-0.4cm}
	\label{fig:limitations}
\end{figure}

We show some failure cases in Fig.~\ref{fig:limitations},
In fact, these failure cases are mainly induced for three main reasons: 1) the binary classifier is not always correct, 2) the patchwise refiner can produce some missing detections, and 3) the proposed key frame strategy might include some low-quality $FS$ in the online fine-tuning process.
One possible method for improvement might be to fully implement our approach in an end-to-end manner, which deserves our future investigation.

The major limitation of our approach might be its offline data mining methodology, which includes multiple sequential steps needing to use multiple off-the-shelf tools such as optical flow tools, object detectors, and ISOD models.
Since our approach is not an end-to-end model, our method is quite time consuming in essence, and we detailed the time cost of each major step in Table~\ref{tab:time}.
Compared with the most representative SSAV19 with 20 FPS (\textbf{f}rame \textbf{p}er \textbf{s}econd), the FPS rate of our approach is only 2.39.

\section{Conclusion and Future Work}
In this paper, we devised a novel long-term scheme for the VSOD task.
The proposed approach can iteratively mine salient object proposals.
The major highlight of the proposed approach is that it converts the conventional framewise VSOD task into an object-level data mining problem.
Moreover, we provided an in-depth analysis and discussion of the rationale of the proposed long-term scheme, which has the potential to benefit our research community in the future.
Additionally, we conducted an extensive component evaluation to verify the effectiveness of each major component in our approach.
Quantitative comparisons to the SOTA models also demonstrate the superiority of the proposed approach.

In the near future, we are particularly interested in reimplementing our approach in a fully end-to-end manner. Thus, the time-consuming limitation can be alleviated, and some empirical parameter settings can also be avoided to make our method more robust.

\vspace{0.4cm}
\textbf{Acknowledgments.} This research was supported in part
by the National Natural Science Foundation of China (62172246), the Open Project Program of State Key Laboratory of Virtual Reality Technology and Systems (VRLAB2021A05), and Youth Innovation and Technology Support Plan of Colleges and Universities in Shandong Province (2021KJ062).

\vspace{-0.1cm}
\bibliographystyle{IEEEtran}
\bibliography{TIP_reference}

\end{document}